\documentclass{acmart}

\usepackage{url}
\usepackage{booktabs}

\usepackage{color}

\begin{document}

\title{Early Detection of Social Media Hoaxes at Scale}

\author{Arkaitz Zubiaga}
\email{a.zubiaga@qmul.ac.uk}
\orcid{0000-0003-4583-3623}
\affiliation{%
  \institution{Queen Mary University of London}
  \city{London}
  \country{United Kingdom}
}

\author{Aiqi Jiang}
\email{a.jiang@qmul.ac.uk}
\orcid{}
\affiliation{%
  \institution{Queen Mary University of London}
  \city{London}
  \country{United Kingdom}
}

\renewcommand{\shortauthors}{Zubiaga, A. and Jiang, A.}

\begin{abstract}
 The unmoderated nature of social media enables the diffusion of hoaxes, which in turn jeopardises the credibility of information gathered from social media platforms. Existing research on automated detection of hoaxes has the limitation of using relatively small datasets, owing to the difficulty of getting labelled data. This in turn has limited research exploring early detection of hoaxes as well as exploring other factors such as the effect of the size of the training data or the use of sliding windows. To mitigate this problem, we introduce a semi-automated method that leverages the Wikidata knowledge base to build large-scale datasets for veracity classification, focusing on celebrity death reports. This enables us to create a dataset with 4,007 reports including over 13 million tweets, 15\% of which are fake. Experiments using class-specific representations of word embeddings show that we can achieve F1 scores nearing 72\% within 10 minutes of the first tweet being posted when we expand the size of the training data following our semi-automated means. Our dataset represents a realistic scenario with a real distribution of true, commemorative and false stories, which we release for further use as a benchmark in future research.
\end{abstract}

%
%

\maketitle

\section{Introduction}

Social media platforms such as Twitter, Instagram and Facebook are increasingly being used by the general public to follow the latest news \cite{sankaranarayanan2009twitterstand,kwak2010twitter} and by journalists for newsgathering \cite{diakopoulos2012finding,gerbaudo2018tweets,zubiaga2013curating,zubiaga2019mining}. 
The fact that anyone can post and share content in social media without moderation enables decentralised production of citizen journalism with an unprecedented detail of report \citep{bruns2012blogs}. However, the unmoderated nature of social media also leads to the production and diffusion of hoaxes \cite{menczer2016spread,allcott2017social}, which exacerbates the credibility of social media as a source for news consumption. With as many as 62\% of citizens using social media for news consumption in 2016 in the US \cite{gottfried2016news}, verification is becoming increasingly important to avoid the spread of misinformation \cite{hermida2012tweets}. This problem is producing an increasing interest in the scientific community to develop automated systems that can determine the accuracy of social media reports with the aim of getting rid of hoaxes \cite{zubiaga2018detection,yavary2020information,alkhodair2020detecting}.

Research in automated detection of misinformation in social media has indeed increased in recent years \cite{shu2017fake,zubiaga2018detection,zhang2020overview}. Researchers have assessed the capacity of average people to identify reports that are inaccurate, finding that their performance leaves much to be desired \citep{kumar2016disinformation}. This reinforces the need to develop automated systems for disinformation detection, however existing work has largely limited to post-hoc classification of reports as true or false, i.e. aggregating an entire timeline of tweets. This means that reports can only be classified hours or even days after they are first released. Research in performing early classification of reports by their truth value is very scarce, partly because of the limited availability of large-scale datasets for the task. An important challenge that hinders the development of early hoax detection systems is the dearth of suitable datasets. Datasets are usually produced by first identifying lists of fake reports. These are then completed by including news reports from other sources to have balanced datasets with fake and real news reports. This, however, is not necessarily representative of a real scenario of incoming reports. This work aims to overcome this issue by introducing a novel approach for generating a large-scale, representative dataset with accurate reports and hoaxes.

In this work we focus on the diffusion of hoaxes and accurate reports of the same type, in an effort to build a representative dataset of accurate and inaccurate reports. A hoax can be defined as a fabricated story that intends to deceive others, such as reporting the existence of a bomb to the police when there isn't one. The word `hoax' originates from the Latin verb \textit{hocus}, meaning ``to cheat'' \cite{nares1822glossary}. In the scientific literature, a `hoax' has been defined by MacDougall as `a deliberately concocted untruth made to masquerade as truth' \cite{macdougall1958hoaxes}.

To develop a representative data collection process collecting hoaxes and a set of comparable truthful reports, we look into death reports of celebrities circulating in social media. Death reports are known to be riddled with hoaxes,\footnote{http://www.snopes.com/tag/celebrity-death-hoaxes/} users frequently making up the death of celebrities, making them viral as if they were real reports and ultimately deceving others. We match these death reports in social media with the entry of the person in question in the Wikidata knowledge base \cite{vrandevcic2014wikidata}. While we conduct experiments online to assess the effectiveness of our models for early detection of hoaxes, our methodology enables performing offline annotations of the data, building a static dataset which we then test simulating a streaming scenario. The advantage of performing the annotation work offline is that we can determine with confidence whether the person really died or not, once the veracity value of the story is settled and its entry in Wikidata is up-to-date. This annotation process has the advantage of being a semi-automated procedure, for instance because one can automatically determine if a person died if the death date on Wikidata matches the date of the tweets reporting the death. When there is no match, it requires more careful manual analysis to determine if it is a hoax, and hence the semi-automated nature of the task, rather than full automation. This semi-automated dataset generation process enables us to create a large-scale dataset with 4,007 death reports over the course of three years (which have over 13 million tweets associated). This dataset can then be exploited in a streaming scenario to determine the earliness with which our system can make accurate predictions on the veracity of the death reports.

In this paper we make the following key contributions:

\begin{itemize}
 \item We propose a novel semi-automated method that leverages the Wikidata knowledge base to build a large-scale dataset for early detection of hoaxes in social media.
 \item We perform experiments using class-specific representations of word embeddings for effective detection of hoaxes. This approach is possible thanks to the semi-automated approach for generation of large-scale datasets, which enables large sets of training data to be available for training the word embedding models of our classifier.
 \item We broaden our set of experiments by looking into the impact of the size of the training data on the classifier's performance.
 \item We look into the use of sliding windows which enables us to leverage the most recent tweets in the timeline associated with a report, instead of the entire timeline. This is motivated by the hypothesis that social media users may exhibit a self-correcting behaviour in these situations, where users may change their mind over time as new reports come out.
 \item We perform an analysis of the social features used in our experiments, which provides insights into the diffusion of death reports, with a focus on distinguishing between hoaxes and accurate reports.
\end{itemize}

Our experimentation shows the effectiveness of our proposed approach for building class-specific word representations, achieving F1 scores of 72\% within just 10 minutes of the first report being posted, and outperforming other baselines. Our experiments also show that the use of sliding windows does not help improve the results; instead, the entire stream of tweets available at the time of classification leads to substantially better results than restricting it to sliding windows of the most recent tweets, hence not validating our hypothesis of a self-correcting behaviour. We also observe that it is important to have a reasonably sized training set to achieve competitive results, with results beginning to plateau only when more than 21 months' worth of data is used for training.

The release of our dataset and trained word embedding models further enable research in veracity classification using a benchmark scenario.

\section{Related Work}

\subsection{Veracity Classification}

Research in determining the accuracy of reports in social media has focused on two different directions: classifying the perceived credibility level of information in social media \citep{castillo2011information,morris2012tweeting,gupta2014tweetcred,shao2018spread,wang2018eann}, and classifying the veracity of social media reports into one of true or false \citep{zubiaga2018detection,shu2017fake,reis2019supervised,tschiatschek2018fake,zhou2019fake}. While the objective in the former is to try to determine the subjective perception of the veracity of a report by its recipients, our objective here aligns with that of the latter, i.e. determining the objective veracity of reports. This is very important for social media users, as it can help flag reports that are classified as fake, as well as to validate reports found to be accurate. Assistance with verification of information in social media is key as previous research found that social media users struggle to identify when information is false \citep{kumar2016disinformation,zubiaga2014tweet}.

Previous work on veracity classification has used different social media platforms including Twitter \citep{takahashi2012rumor} and Sina Weibo \citep{yang2012automatic}. However, most of this work has performed post-hoc classification of reports as true or false \citep{jin2016news,ma2016detecting,tacchini2017some}, which means that they need to observe the entire development of a story before classifying it. This may imply hours or even days of delay by the time a story can be classified. Our objective here instead is to aim for early classification of stories, with the ultimate goal of detecting hoaxes early on.

Research looking into either real-time or early detection of hoaxes is scarce. \cite{liu2015real} use a set of features including user metadata and propagation structure to verify stories within hours of being posted for the first time. They show competitive performance with the use of both feature sets 72 hours after the story was first posted. Another approach is presented by \cite{sampson2016leveraging}, combining hashtags and links as features to determine the veracity of reports. They report results between 1 and 10 hours, with results increasingly improving over time. While both of these are clever approaches that are worthwhile considering, neither of their systems was publicly released and the features used in their experiments are not reproducible with the level of detail provided.

There is also work tackling early detection of ``fake news.'' \cite{liu2018early} define the early detection task as that consisting in determining the truth value of a single tweet. They look at the propagation of the tweet through retweets, using the user profiles to try to determine early on the veracity value of the tweet. This method is limited to retweets of a single tweet, hence the authors only look at features from the profiles of those retweeting; this differs from our study where we consider multiple tweets associated with each story (i.e. a death report), and therefore our experiments look at the capacity of determining the veracity of a story by aggregating related tweets over time in a streaming scenario. In \cite{zhou2019fake2}, the authors used a dataset of tweets with associated fact-checks from professional organisations. This is also a sensible methodology, to which our methodology contributes by defining a novel methodology to come up with a large-scale dataset with annotations grounded in associated Wikipedia pages, for a type of story in which the veracity can easily be determined with confidence post-hoc, celebrity death reports. In other cases, researchers have referred to their task as ``early detection of fake news,'' as is the case with Gereme et al. \cite{gereme2019early}, it is however unclear how the temporal aspect of the stories has been incorporated into their models, as it is never discussed.

Others have taken a different approach by using stance classifiers \citep{qazvinian2011rumor,derczynski2017semeval,gorrell2018rumoureval,zubiaga2018discourse,zubiaga2016stance,dungs2018can}. Instead of using a classifier that directly outputs one of true or false given a report as input, they try to determine the stance that each social media post expresses with respect to a report, such as supporting, denying, querying or commenting. They then propose to aggregate the different stances to determine the likely veracity of a report. While this is a sensible approach, it also requires a large amount of posts to be observed in order to aggregate the different stances, which may impede early determination of report veracity.

Research in early detection of veracity in social media is still limited, largely hindered by the lack of suitable datasets that enable experimentation in a streaming scenario. Our benchmark dataset and experimentation aims to fill this gap.

\subsection{Related Datasets}

Research in veracity classification has been largely limited by the dearth of proper datasets. This is changing in recent years, however often with limitations in representativity of the dataset contents or quality of veracity annotations. As \cite{shu2017fake} stated, development of a dataset annotated for veracity is very challenging, as judgments from professionals are generally needed to carefully verify and subsequently annotated stories. As shown by previous research \citep{kumar2016disinformation}, average users struggle to distinguish true and false stories. It is therefore not generally a suitable task to be performed through crowdsourcing, requiring careful analysis of stories either through professional input or by checking reputable sources or evidence. As a result, few representative datasets have been produced. Most of these datasets are created by first collecting false stories, and then completing the datasets with randomly picked true stories \citep{kwon2017rumor,liu2015real}. The use of different methodologies for collecting false and true stories is however not ideal as it will inevitably differ from a real scenario. Furthermore, existing datasets are normally made of isolated posts annotated for veracity (cf. \cite{wang2017liar,tacchini2017some}), which pose limitations when one wants to investigate the earliness of veracity classification models in a incoming stream of multiple posts linked to a single story. To test our models at different points in time on a streaming scenario, we need to collect a timeline of tweets linked to each story instead.

Recent years have seen a surge of datasets to research in the misinformation landscape, dominated mostly by those containing isolated posts and hence not enabling research in early detection. The vast majority of these datasets are made of fact-checks collected from professional organisations, with claims or headlines labelled as true or false (or a wider spectrum with combinations of these, such as mostly true, mostly false and half true). This is the case of NELA-GT-2018 \cite{norregaard2019nela} and FakeNewsNet \cite{shu2018fakenewsnet} in English, GermanFakeNC \cite{vogel2019fake} in German and Factck.br \cite{moreno2019factck} in Portuguese. These can be deemed high quality annotations, particularly when they are collected from fact-checking organisations recognised by the International Fact-Checking Network (IFCN),\footnote{\url{https://www.poynter.org/ifcn/}} however their representativity can be questionable as it is dependent on the editorial selection of stories by the fact-checking organisation in question. Another dataset for fact-checking claims is FEVER \cite{thorne2018fever}, which collected and altered claims extracted from Wikipedia, automatically creating correct and incorrect claims for fact-checking; this is a clever approach to create a large-scale dataset, which however does not enable exploration of earliness in classification due to claims being isolated. Others have relied on the quality of sites to determine if their news articles are real or fake, i.e. a collection of articles from a reputable news organisation (e.g. Wall Street Journal) would be deemed accurate, whereas articles from low quality or parody news outlets (e.g. The Onion) would be deemed fake (cf. FakevsSatire \cite{golbeck2018fake}, FA-KES \cite{salem2019fa}, Newsbag \cite{jindal2019newsbag}); this enables easy collection of large-scale datasets, however it raises concerns about the quality of the annotation, as well as whether the final classification task consists in determining the veracity of articles or instead in classifying the source of the news.

r/Fakeddit \cite{nakamura2019r} provides a large-scale, representative collection of Reddit posts, where labels are however automatically determined by using machine learning models, which cannot guarantee high quality of labels. Credbank \cite{mitra2015credbank} is another related dataset, which however includes annotations for perceived credibility scores, rather than actual veracity scores.

Work on the PHEME project \cite{zubiaga2016analysing} focused instead on rumours, i.e. stories that start off as unverified. Through the organisation of two shared tasks, RumourEval \cite{derczynski2017semeval,gorrell2018rumoureval}, the project looked at how the stances expressed by users over time can help determine the veracity of rumours early on. This is one of the most related datasets to the present work, which however does not scale as easily as it required manual input from journalists to determine the veracity of rumours. The data collection and annotation approach defined in this work is semi-automated, enabling generation of large-scale datasets.

In this work, we describe a novel approach for semi-automated dataset generation, which removes the sampling bias as verification of larger sets of instances is possible through the use of Wikidata as an external source. Likewise, our approach enables collections of both true and false stories by following the same methodology, leading to the first large-scale, representative dataset collected out of social media.

\subsection{Learning Class-specific Word Representations}

Class-specific word representations have been found to be useful for different classification tasks, as is the case with the use of Brown clusters to build class-specific language models \citep{brown1992class}. Brown clusters have been successfully used by researchers for training word representations \citep{turian2010word}, natural language processing tasks such as dependency parsing \citep{koo2008simple} or for building class-specific language models \citep{bengio2003neural}, among others. As a state-of-the-art approach for semantic word representation, here we make use of word embeddings \citep{mikolov2013distributed}. We propose to train and leverage class-specific word embeddings to learn the patterns of each class in the training data. The difficulty to achieve this generally lies in the necessity for large-scale annotated datasets that have large numbers of instances for each class. Our semi-automated approach for building large-scale annotated datasets enables to have large collections of data to train class-specific word embeddings.

\section{Materials and Methods}

\subsection{Dataset}

Our data collection methodology is semi-automated, involving little and easy human input, which enabled us to collect a large-scale dataset. The dataset generation process consists of three steps: (1) data collection, (2) linking to Wikidata, and (3) data annotation.

\subsubsection{Data collection}

We first perform keyword-based collection of tweets from Twitter. We use `RIP' as a keyword that is largely associated with death reports. Twitter's results are not case sensitive, so we collect all tweets including the keyword and remove those that are not upper-cased at a later stage. We perform the collection of tweets containing the keyword `RIP' for a period of three years between January 1, 2012 and December 31, 2014. This longitudinal data collection led to a total of over 94.2 million tweets.

\subsubsection{Linking to Wikidata}

As we completed the collection of tweets at the end of 2014, we downloaded a dump of Wikidata \cite{vrandevcic2014wikidata} in January 2015, which is a structured knowledge base that includes, among others, an extensive database of notable people, in part extracted from Wikipedia but also completed by volunteer contributors. The entries of these notable people in the knowledge base include their death date, when the person deceased; a null value as the death date indicates the person is alive. We used its API to download all entries corresponding to people,\footnote{To identify entries that are about people, we looked for entries with the property ``P569'', which refers to ``date of birth'' and is therefore indicative of an entry belonging to a person: \url{https://www.wikidata.org/wiki/Property:P569}} leading to a collection of 1,136,543 different people. Each of these entries includes the fields shown in the following example:

\begin{description}
 \item
 ~ \newline
 \footnotesize
 \texttt{\{"id":"8023",}\\
 \texttt{"name":"Nelson Mandela",}\\
 \texttt{"birth":\{"date":"1918-07-18","precision":11\},}\\
 \texttt{"death":\{"date":"2013-12-05","precision":11\},}\\
 \texttt{"description":"former President of South Africa, anti-apartheid activist",}\\
 \texttt{"aliases":["Nelson Rolihlahla Mandela","Mandela","Madiba"]\}}
 \label{ex:wikidata-entry}
\end{description}

We are interested in most of these features for our research, but especially in the name and aliases, which we use to identify mentions of people in our `RIP' tweets, and also the death date, which indicates if a person is still alive or has died on a particular date. Note that birth and death dates have a precision value associated, which refers to the granularity of the date. A value of 11 implies the date is accurate at the day level. The standard for contemporary people is for this value to be 11. Year and month-level precision scores are occasionally given for people in earlier centuries. We use the Wikidata knowledge base to look for mentions of contemporary people in our Twitter dataset, and so the lack of precision for ancient people does not have an effect in our case.

Having the collection of `RIP' tweets and the entries for people on Wikidata, we look within the tweets for mentions of names (and aliases) of people in the Wikidata knowledge base, e.g. tweets containing `RIP Nelson Mandela'. To do so, as a first step, since the keyword search on Twitter is case insensitive, we removed all occurrences where the keyword `RIP' was not completely upper-cased. We then looked for tweets where the keyword `RIP' was followed by one of the person names (or aliases) in Wikidata. We do this for all the tweets and keep the instances in which the name of a person is mentioned at least 50 times in a day. Removing instances with fewer than 50 tweets reduces noise from spam tweets that did not go viral, and makes the manual annotation (which we explain below) more manageable. Note that this process can also identify numerous instances of mentions of the same person, i.e., being reported dead in social media more than once within the time frame of our study between 2012 and 2014. Consecutive days mentioning the same person are considered part of the same death instance, while we only consider a new instance when there is at least one day gap between mentions. This process led to a dataset with 4,007 death reports pertaining to 3,066 different people. The total number of tweets associated with these reports amounts to 13,302,600.

\subsubsection{Description of the Hoax Detection Task}

The hoax detection task consists in identifying emerging reports that are false. In our experiments, we aim to identify the death reports that have been fabricated, i.e. reporting cases of deaths that have not actually happened. We formally define the death hoax detection task as that in which a supervised classifier has to determine which of the following three categories a new incoming reporting belongs to: $Y = \{real, commemoration, fake\}$. We use three categories as we distinguish cases of \textit{fake} reports, where a death has been fabricated, \textit{real} reports, where a death report has indeed recently happened, and \textit{commemorations}, where a past death is being remembered. In what follows we detail the annotation process we relied on.

\subsubsection{Annotation}

At this stage we have 4,007 death reports linked to Wikidata pages. To conduct the annotation of these death reports, we developed an annotation tool that visualises the stream of tweets associated with a report, along with a form that enables the annotation. Tables \ref{tab:real-death}, \ref{tab:commemoration-death} and \ref{tab:fake-death} show the information we provide in the annotation tool, with three examples for real, commemorative and fake death reports.

\begin{table}
 \footnotesize
 \begin{tabular}{| l p{1cm} l |}
  \hline
  \textbf{Death report on: 12th December, 2014} &  & \textbf{Wikidata entries} \\
   &  & \#1: personname (death: 12-12-2014, born: 1940) \\
  RIP personname ... &  & \#2: personname  (death: 0, born: 1975) \\
  RIP personname ... &  &  \\
  RIP personname ... &  & $\bigotimes$ Real \\
  RIP personname ... &  & $\bigcirc$ Commemoration \\
   &  & $\bigcirc$ Fake \\
  \hline
 \end{tabular}
 \caption{Example of real death report, where the date of the death report and the death date of a Wikidata entry match. Note there are two Wikidata entries matching the person name in question in this case, where the death date of one of them matches that of the death report.}
 \label{tab:real-death}
 
 \begin{tabular}{| l p{1cm} l |}
  \hline
  \textbf{Death report on: 12th December, 2014} &  & \textbf{Wikidata entries} \\
   &  & personname (death: 12-12-2009, born: 1945) \\
  RIP personname ... &  &  \\
  RIP personname ... &  & $\bigcirc$ Real \\
  RIP personname ... &  & $\bigotimes$ Commemoration \\
  RIP personname ... &  & $\bigcirc$ Fake \\
  \hline
 \end{tabular}
 \caption{Example of a commemorative death report, where the date of the death report and the death date of a Wikidata entry are exactly years apart from each other, hence indicating that Twitter users are remembering the person who died years ago.}
 \label{tab:commemoration-death}
 
 \begin{tabular}{| l p{1cm} l |}
  \hline
  \textbf{Death report on: 12th December, 2014} &  & \textbf{Wikidata entries} \\
   &  & personname (death: 0, born: 1972) \\
  RIP personname ... &  &  \\
  RIP personname ... &  & $\bigcirc$ Real \\
  RIP personname ... &  & $\bigcirc$ Commemoration \\
  RIP personname ... &  & $\bigotimes$ Fake \\
  \hline
 \end{tabular}
 \caption{Example of fake death report, where the matching Wikidata entry has no death date (i.e. death date = 0).}
 \label{tab:fake-death}

\end{table}

Most importantly, the annotation tool shows the date in which the death report broke on Twitter, along with a list of Wikidata entries of candidates matching the person mentioned in the tweets. An exact match between the date of the death report and the death date of one of the Wikidata entries is then highly indicative of a real death report linked to that candidate. Hence, for a majority of the death reports, a tentative annotation candidate can be done automatically by the tool, in the following cases:

\begin{itemize}
 \item If the date of the death report and the death date of one of the Wikidata entries match, the annotation tool will automatically mark the death as being real. Note that besides exact date matches, we also automatically mark it as a real death if the date of the report and the death date of a Wikidata entry are only one day apart, due to time zone differences (i.e. tweets being UTC and the person dying elsewhere in the world).
 \item If the date of the death report and the death date of a Wikidata entry match (or they are one day apart) but on a different year, then we automatically mark it as a commemoration.
 \item If there is a single Wikidata entry listed as a candidate and that entry has not died (death date = 0), then we mark it as fake.
\end{itemize}

These automated annotations are then shown to the annotator, who supervises and approves (or changes) the annotation, which is substantially faster than annotating them from scratch. For the cases that do not match any of the conditions listed above, the annotation is done from scratch.

The annotation process is done post-hoc (not in real-time), and therefore Wikidata entries were collected much later, after the whole three years comprised in the dataset were collected. This avoids potential cases of wrong updates on Wikidata impacted by fake reports on Twitter.

\subsubsection{Final Dataset}

The annotation of the 4,007 death reports in our dataset led to the following distribution: 2,301 real deaths, 1,092 commemorations and 614 fake deaths. Table \ref{tab:dataset} shows the statistics of the dataset. While the categories are imbalanced, this still shows that fake deaths represent a large proportion of all reports (15.3\%) and need to be tackled to avoid their diffusion. The skewed distribution of categories presents in turn an additional challenge for the classification task.

The annotation was primarily done by a single annotator, by supervising the automated annotation of the tool described above. The validate the quality of the annotation, a second annotator went through a random subset of 200 death reports, achieving a high inter-rater agreement measured with a Cohen's Kappa of 0.982 \cite{cohen1960coefficient}.

\begin{table}[htb]
 \centering
 \begin{tabular}{c c c}
  \toprule
  \textbf{Veracity} & \textbf{Instances} & \textbf{Tweets} \\
  \midrule
  Real & 2,301 & 9,131,976 \\
  \midrule
  Commemoration & 1,092 & 526,588 \\
  \midrule
  Fake & 614 & 643,432 \\
  \midrule
  Total & 4,007 & 10,301,996 \\
  \bottomrule
 \end{tabular}
 \caption{Distribution of labels and tweets in the dataset.}
 \label{tab:dataset}
\end{table}

\begin{figure}
  \begin{center}
   \includegraphics[width=0.45\textwidth]{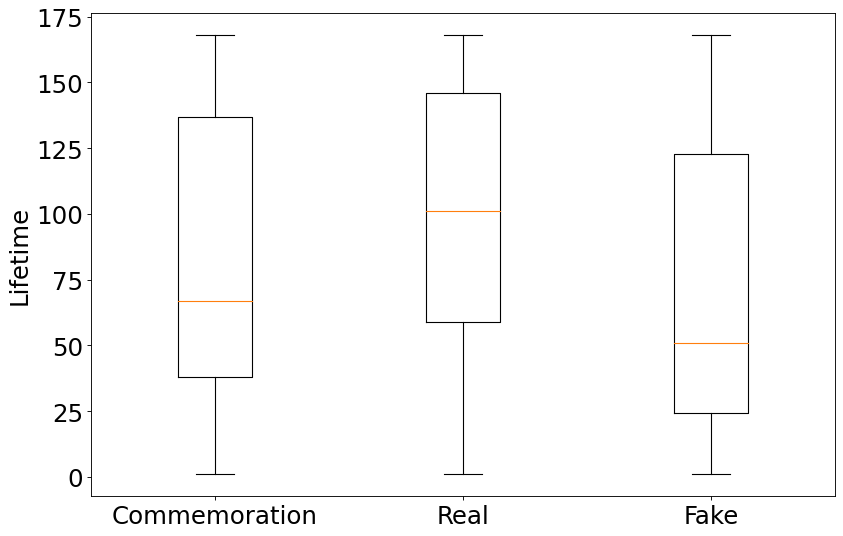}
   \Description{Lifetimes of death reports grouped by type: commemorations, real deaths and fake deaths.}
   \caption{Lifetimes of death reports grouped by type: commemorations, real deaths and fake deaths.}
   \label{fig:lifetimes}
  \end{center}
 \end{figure}

\begin{figure}
  \begin{center}
   \includegraphics[width=0.45\textwidth]{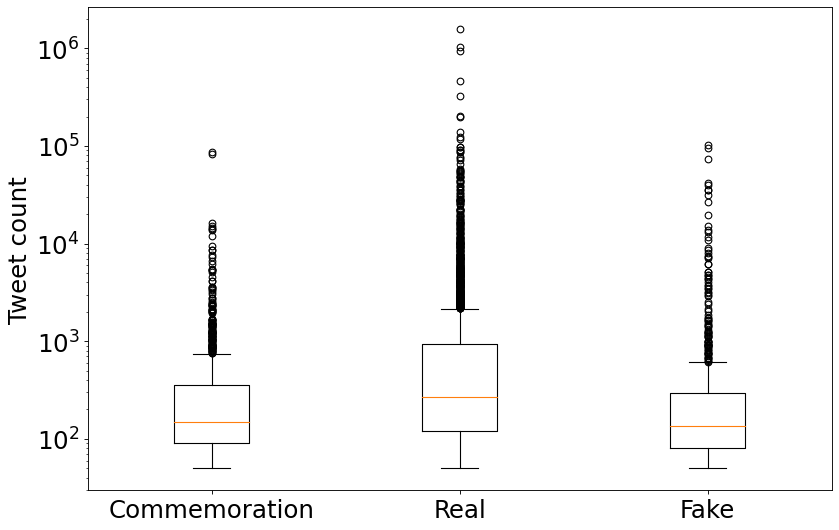}
   \Description{Tweet counts of death reports grouped by type: commemorations, real deaths and fake deaths.}
   \caption{Tweet counts of death reports grouped by type: commemorations, real deaths and fake deaths.}
   \label{fig:tweetcounts}
  \end{center}
 \end{figure}

We look at the lifetime (Figure \ref{fig:lifetimes}) and the number of tweets (Figure \ref{fig:tweetcounts}) of different kinds of death reports.\footnote{Note that we only preserve up to 7 days of tweets associated with a death report, as we do not need more for the purposes of our research on early detection of hoaxes. Death reports that may have lasted longer are therefore truncated to 168 hours (7 days) in Figure \ref{fig:lifetimes}.} Interestingly, we can see that fake reports have a tendency to last shorter and have fewer tweets posted; still, the median fake reports lasts for about 50 hours. This can be of interest for a behavioural analysis comparing hoaxes and real reports, however it is of little help for an early hoax detection system if we aim to detect hoaxes within a short time after first being posted. Commemorations also have a tendency to last shorter than real deaths, perhaps understandably as the emotional impact of a commemoration is expect to be lower than that of a recent death.

The death reports with the highest number of tweets posted, by accuracy of report, include:
\begin{itemize}
 \item \textbf{Real deaths:} Robin Williams (1.58M tweets), Paul Walker (1.04M), Nelson Mandela (939K), Whitney Houston (462K), Neil Armstrong (324K), Maya Angelou (203K), Cory Monteith (198K), Casey Kasem (139K), Philip Seymour Hoffman (122K), Jenni Rivera (118K).
 \item \textbf{Commemorations:} James Avery (86K tweets), John Lennon (83K), Kurt Cobain (16K), Steve Irwin (15K), Jesus Christ (15K), Eric Garner (14K), Sean Taylor (14K), Gary Speed (12K), George Best (9K), Helen Martin (8K).
 \item \textbf{Death hoaxes:} Megan Fox (101K tweets), Lady Gaga (95K), Chris Brown (73K), Margaret Thatcher (53K), Taylor Swift (40K), Justin Bieber (36K), Eddie Murphy (35K), Channing Tatum (32K), Rowan Atkinson (27K), Ricardo Arjona (20K).
\end{itemize}

The people who were most repeatedly killed off through death hoaxes include: Justin Bieber (13 times), Soulja Boy (11), Chris Brown (10), Lady Gaga (9), Nicki Minaj (9), Taylor Swift (8), 50 Cent (7), Chuck Norris (7), Eddie Murphy (7), Lebron James (7).

The following are examples of tweets associated with a death hoax.

\begin{description}
 \item
 ~ \newline
 \footnotesize
 \texttt{\#1: RIP Cesar Millan \#ripcesarmillan \#dogwhisperer}\\
 \texttt{\#2: RIP CESAR MILLAN - You did so much wunnerful fings fah doggiez!}\\
 \texttt{\#3: RIP Cesar Millan. I'm sure u will be missed by alot of people, especially the dog lovers}\\
 \texttt{\#4: RIP Cesar Millan, gone to soon. Hope you run free over rainbow bridge with all the dogs. \#dogwhisperer}\\
 \texttt{\#5: We will miss you Cesar we love your show RIP Cesar Millan}\\
 \texttt{\#6: So young!! Ugh. RIP Cesar Millan, aka The Dog Whisperer. Very Sad.}\\
 \label{ex:death-hoax}
\end{description}

The hoax went viral with tweets originating from multiple sources and with tweets with very different textual content. Later tweets in the timeline of this viral timeline of tweets start warning others that the story is a hoax: \textit{It's hoax, eat that!}. Despite later correction by Twitter users, the story went viral in what one may consider a credible story, especially if they do not check additional sources for verification.

\subsection{Hoax Detection}

In this section we provide details of the features and experiment settings that we use for our work.

\subsubsection{Classification Features}

We use three different types of features, including two features that are widely used in previous work (social features and textual features), as well as our proposed class-specific word representations. Additionally, we propose two different combinations of those features. To simulate the task of early detection of hoaxes, we perform experiments at different points in time. Experiments performed in time $t$ will generate the features only from tweets posted before that time. The feature sets we use for the experiments are as follows:

\begin{itemize}
 \item \textbf{Baseline 1 -- Social features (social):} We use a set of 16 features that refer to the reputation of the users participating in a report and to diffusion patterns. Please see the appendix for more details of these features.
 \item \textbf{Baseline 2 -- Textual features using word embeddings (w2v):} As a state-of-the-art word representation approach, we use Word2Vec embeddings \citep{mikolov2013distributed} to represent the content of the tweets associated with a report. The model we use for the embeddings was trained from the entire collection of tweets in the training set, i.e. all the 2012 and 2013 tweets. We represent each tweet as the average of the embeddings for each word, and finally get the average of all tweets.
 \item \textbf{Baseline 3 -- Google's word embedding model (gw2v):} As one of the standard and most commonly used word embedding models, we use Google's Word2Vec model with 300 dimensions trained from Google News data\footnote{\url{https://code.google.com/archive/p/word2vec/}} as a baseline for comparison with our model.
 \item \textbf{Baseline 4 -- Textual features using sentence embeddings (infersent):} InferSent \cite{conneau2017supervised} is a state-of-the-art method for semantic representation of sentences into embeddings. Beyond word embeddings, which consider each word as a separate element and ignore aspects like the order and importance of words in a sentence, sentence embeddings methods like InferSent can capture syntactic features. InferSent is a method that learns the structure of sentences from a large natural language inference corpus, the Stanford Natural Language Inference (SNLI) corpus \citep{bowman2015large}, by using a Bi-LSTM encoder.
 \item \textbf{Class-specific word representations (multiw2v):} The same word can have different meanings depending on the category in which it is used. For instance, `RIP' usually refers to `Rest In Peace' or `Requiescat In Pace' when it is used along with a real death, but it can mean `Really Inspiring Person' when used as a hoax; this is however implicit.\footnote{Note that tweets using the `RIP' keyword all look the same. It is only later when users clarify that they were joking and using the keyword `RIP' to refer to `Really Inspiring Person'. Here we mean that class-specific word embeddings can model the keyword `RIP' differently for real and fake deaths if their context varies.} This can be hard to distinguish even for humans as the word is exactly the same, but it can be modelled differently using class-specific word embeddings. Provided that we have large-scale training data, we propose to train different word embedding models for each class, so that each model learns the vocabulary of that class. We build three different collections from our training set, each belonging to tweets from one of the categories, and train a separate word embedding model from each of the three collection, so that we have a word embedding for \textit{real} reports, another one for \textit{fake} reports and a third one for \textit{commemorating} reports. Having three different word embedding models (real, fake, commemoration), we then create three different vectors, each of which is created as above, however using a different word embedding model. Finally, we combine all three vectors by concatenating them into a single vector. Our proposed model, which we call \textit{multiw2v}, enables characterisation of reports with respect to each class in the dataset.
\end{itemize}

We also test combinations of social and different textual features, including word embeddings (social+w2v), sentence embeddings (social+infersent) and class-specific word representations (social+multiw2v).

\subsubsection{Experiment Settings}

Given that the objective of our experimentation is to find out what features perform best for early detection of hoaxes, assessing the performance of our proposed class-specific word representations, we first tested different classifiers: Support Vector Machines, Random Forests, Logistic Regression, Multi-layer Perceptron, Gaussian Processes and Naive Bayes. We found the Logistic Regression classifier \citep{ratnaparkhi1997simple} to perform substantially better than the rest of the classifiers, and so for the sake of clarity and space we show results for this classifier in the rest of this article. Additional results for the rest of the classifiers are given in the appendices. We use the implementation of the logistic regression classifier in scikit-learn \citep{pedregosa2011scikit},\footnote{\url{http://scikit-learn.org/}. We use scikit-learn 0.22.2 on Python 3.6.3.} with the following parameters:

\begin{description}
  \item
  \texttt{solver='liblinear', multi\_class='ovr', fit\_intercept=True,} \\
  \texttt{intercept\_scaling=0.0001, C=0.6, class\_weight='balanced'}
\end{description}

These parameters were determined empirically by testing a range of different parameters on 10-fold cross-validation experiments on held-out parts of the training data. We tested all possible values for categorical parameters, whereas a wide range of values were tested for the numerical parameters, keeping the best-performing parameters in each case and re-testing with nearby values. The high performance computing infrastructure provided by our university was employed for all the experiments.\footnote{\url{https://docs.hpc.qmul.ac.uk/}}

For the experimentation, We use the first two years (2012 and 2013) for training and the last year (2014) for testing. With this we avoid mixing data from overlapping periods in the training and test sets, and also other cases like having newer data in the training set than in the test set (e.g. if a person died in 2014, we avoid having the real 2014 death in the training set and a fake 2013 death in the test set)\footnote{This happened for instance with Nelson Mandela, who was killed off multiple times throughout 2012 and 2013 before he actually died in December 2013}. In addition, using old data for training and new data for testing allows simulating a more realistic scenario. Despite having static sets for training and test, we run 10-fold cross-validation experiments with different subsets of the training data. We opted for doing 10-fold cross-validation to enable generalisation of the results and to avoid skewed results affected by a specific training set.

We report performance scores of different classifiers using macroaveraged F1 scores, i.e. averaged F1 scores for the three categories, where the F1 score for a category equates to the harmonic mean between the precision and the recall.

\section{Results and Discussion}

We first present a comparison of the different features under study, delving into results by category. Then, we explore the use of sliding windows for the classification.

\subsection{Comparison of Features}

We first compare the sets of features and combinations of features we described above. We show results for classification experiments in different points in time including 0 (only the first tweet posted), 5, 10, 15, 30, 60, 120, 180 and 300 minutes. This allows us to explore the ability to perform accurate classification early on in the first few minutes, as well as to analyse how much the classifier's performance can improve as time goes on up to 5 hours.

Table \ref{tab:features} shows the results comparing performance of different features. We observe that the approaches using our proposed method for class-specific word representations (multiw2v) perform better than the rest, including the use of standard word embeddings (w2v and gw2v) as well as sentence embeddings (infersent). While social features alone perform poorly, they are actually beneficial when they are combined with the multiw2v features. We see that the combination of social+multiw2v consistently outperforms the sole use of multiw2v features, however this improvement is especially noticeable for later points in time, as the social features become more beneficial with more tweets observed over time; i.e. when the social trend develops. For very early detection of hoaxes, both multiw2v and social+multiw2v perform similarly, with a slightly better performance for the latter. While it is possible to have fairly accurate classification having only observed the first tweet (.669), it is worthwhile delaying the prediction for 2 to 10 minutes to achieve an improved performance (0.696 and 0.716). It is only in later stages in the diffusion of hoaxes, after 5 hours, that the combination of social and InferSent features manages to perform slightly better than the social+multiw2v features; this is, however, not ideal for early detection of hoaxes, where we are especially interested in performance results for the early stages of the stream. For earlier stages of the stream, multiw2v-based features clearly outperform InferSent and the other baselines.

\begin{table*}[tbh]
 \begin{center}
  \begin{tabular}{l c c c c c c c c c c}
   \toprule
    & 0 & 1' & 2' & 5' & 10' & 15' & 30' & 60' & 120' & 300' \\
   \midrule
   social & .427 & .495 & .509 & .510 & .510 & .528 & .535 & .577 & .594 & .591 \\
   \hline
   w2v & .641 & \textit{.655} & .658 & .663 & .667 & .670 & .680 & .696 & .699 & .698 \\
   social+w2v & .612 & .634 & .661 & .671 & .671 & .677 & .675 & .709 & .709 & .724 \\
   \hline
   gw2v & .556 & .565 & .574 & .608 & .612 & .618 & .623 & .645 & .648 & .664 \\
   social+gw2v & .569 & .590 & .599 & .616 & .633 & .647 & .663 & .679 & .688 & .686 \\
   \hline
   infersent & .637 & .640 & .653 & .664 & .683 & .681 & .697 & .722 & .734 & .759 \\
   social+infersent & \textit{.643} & \textit{.655} & \textit{.670} & \textit{.678} & \textit{.691} & \textit{.688} & \textit{.698} & \textit{.731} & \textit{.748} & \textbf{.767} \\
   \hline
   multiw2v* & \textbf{.669} & .676 & .691 & .703 & .714 & .722 & .723 & .721 & .738 & .741 \\
   social+multiw2v* & .647 & \textbf{.677}$\ddagger$ & \textbf{.696}$\ddagger$ & \textbf{.707}$\ddagger$ & \textbf{.716}$\ddagger$ & \textbf{.725}$\ddagger$ & \textbf{.724}$\dagger$ & \textbf{.744}$\dagger$ & \textbf{.752} & \textit{.748} \\
   \bottomrule
  \end{tabular}
  \caption{Comparison of features for early detection of hoaxes. Proposed methods indicated with a star (*). Best method highlighted in bold and second best method for different types of features highlighted in italic. $\ddagger$: statistically significant at $p<.01$, $\dagger$: statistically significant at $p<.05$.}
  \label{tab:features}
 \end{center}
\end{table*}

\subsection{Effect of the Size of the Word Embedding Model}

\begin{table*}[tbh]
 \begin{center}
  \begin{tabular}{l c c c c c c c c c c}
   \toprule
    & 0 & 1' & 2' & 5' & 10' & 15' & 30' & 60' & 120' & 300' \\
   \midrule
   w2v300 & .641 & .655 & .658 & .663 & .667 & .670 & \textbf{.680} & \textbf{.696} & \textbf{.699} & .698 \\
   w2v600 & .631 & .643 & .656 & .657 & .658 & .663 & .669 & .681 & .690 & .693 \\
   w2v900 & .640 & .658 & \textbf{.667} & \textbf{.672} & \textbf{.668} & \textbf{.679} & .679 & .695 & \textbf{.699} & \textbf{.706} \\
   w2v4096 & \textbf{.642} & \textbf{.664} & .663 & .667 & .666 & .671 & .677 & .695 & .698 & .697 \\
   \hline
   social+w2v300 & .612 & .634 & .661 & .671 & .671 & .677 & .675 & \textbf{.709} & \textbf{.709} & \textbf{.724} \\
   social+w2v600 & .620 & \textbf{.654} & \textbf{.669} & .675 & .670 & \textbf{.682} & .685 & .707 & .707 & .712 \\
   social+w2v900 & \textbf{.621} & .646 & .660 & \textbf{.680} & \textbf{.672} & .681 & \textbf{.688} & .698 & .705 & .715 \\
   social+w2v4096 & .614 & .649 & \textbf{.669} & .675 & .668 & .681 & .683 & .697 & .702 & .713 \\
   \bottomrule
  \end{tabular}
  \caption{Comparison of performance by using word2vec models of varying dimensionality, including 300 (original w2v above), 600, 900 (equivalent to multiw2v) and 4096 (equivalent to InferSent).}
  \label{tab:word-embedding-size}
 \end{center}
\end{table*}

In an additional set of experiments, we aim to determine the extent to which the number of dimensions in the embedding model can impact performance. The motivation behind this experimentation is that the original word2vec model (w2v) creates vectors with 300 dimensions, class-specific embeddings (multiw2v) create vectors with 900 dimensions, and infersent creates vectors with 4096 dimensions. Hence, we perform additional experiments with word2vec models trained with different dimensionalities, which can help us determine if the improvement achieved by multiw2v comes because of the different methodology used or simply because of the higher dimensionality.

Table \ref{tab:word-embedding-size} shows results for word2vec models of dimensionalities of 300, 600, 900 and 4096, both on their own and combined with social features. We can observe that variation in performance is marginal and shows that higher dimensionality does not lead to improved performance. While this marginal improvement fluctuates slightly, we can observe that on occasions even the use of 300 dimensions can outperform bigger models with 4096 dimensions (e.g. after 10, 30 or 60 minutes). These results demonstrate the potential of multiw2v to provide substantial improvements thanks to leveraging class-specific embeddings and not because of the larger dimensionality.

\subsection{Using Sliding Windows}

We now experiment with the use of sliding windows for the classification \cite{datar2002maintaining}. With sliding windows, we can choose to make use of all the tweets posted so far for a report at time $t$ to classify it, or we can instead make use of a smaller window that only uses the last bit. The motivation behind this is that we hypothesise that Twitter users will show a self-correcting behaviour, potentially being mistaken about the truth of a report in the very early stages, but later correcting themselves as new evidence or more sources are available related to the report. We experiment with different sliding windows by using different percentages. For each percentage, we consider the tweets posted within that fraction of time, counting from the end: $w = \{t - (t - t_0) * p, t\}$, where $w$ is the window comprised between: (1) the current time $t$ minus the percentage $p$ of time between the current time and the time of the first tweet was posted, and (2) the current time.

Table \ref{tab:windows} shows the results of using different time windows: 0.1, 0.25, 0.5, 0.75 and 1.0. We use the \textit{social+multiw2v} as the best performing features here for the analysis. With these results we observe that the use of sliding windows is not useful, and that it is much better to use all the tweets associated with a report than the last few. While we do observe that it is better to keep including new tweets as time goes on, which leads to performance gains, we also see that it is important to include all the tweets from the very beginning. Note that results for $t = 0$ are the same in all cases as the use of a sliding window does not have an effect in this case. These results do not support our hypothesis of a self-correcting behaviour happening among users; while some users may possibly correct themselves, there is no sufficient impact on the model to improve performance, hence not validating our hypothesis. Note, however, that we cannot reject the hypothesis for not having investigated it in detail; our experiments lead to the conclusion that the way we modelled this potential self-correcting behaviour leads to performance drop.

\begin{table*}[tbh]
 \begin{center}
  \begin{tabular}{l c c c c c c c c c c}
   \toprule
   window & 0 & 1' & 2' & 5' & 10' & 15' & 30' & 60' & 120' & 300' \\
   \midrule
   0.1 & \textbf{.647} & .385 & .399 & .413 & .423 & .442 & .452 & .459 & .466 & .514 \\
   0.25 & \textbf{.647} & .422 & .468 & .476 & .478 & .519 & .522 & .547 & .582 & .617 \\
   0.5 & \textbf{.647} & .228 & .284 & .369 & .537 & .544 & .575 & .589 & .642 & .673 \\
   0.75 & \textbf{.647} & .253 & .319 & .396 & .554 & .580 & .598 & .626 & .671 & .718 \\
   1.0 & \textbf{.647} & \textbf{.677} & \textbf{.696} & \textbf{.707} & \textbf{.716} & \textbf{.725} & \textbf{.724} & \textbf{.744} & \textbf{.752} & \textbf{.748} \\
   \bottomrule
  \end{tabular}
  \caption{Results using sliding windows for early detection of hoaxes, using the best performing set of features (social+multiw2v).}
  \label{tab:windows}
 \end{center}
\end{table*}
 
For more results on the impact of sliding windows using other baseline classifiers, please refer to Appendix \ref{ap:baselines}.

\subsection{Effect of the Size of the Training Data}

We now analyse the effect of using different sizes of the training set for the experimentation. We have two years' worth of tweets in the training set, but here we are interested in exploring if we could achieve comparable results by using less training data, which would alleviate the need for having to collect more data prior to running the classifier. We analyse the use of eight different sizes of training sets, with a step size of 3 months between them, i.e. 3, 6, 9, 12, 15, 18, 21 and 24. When we use a number $N$ of months in our training data, we are taking the first $N$ months starting from the beginning of our dataset in January 2012, e.g. 3 months includes January, February and March 2012.

Table \ref{tab:training} shows the results for the use of training sets of different sizes. We observe substantial improvements for the smaller numbers of months, but these improvements become much smaller as we have more training data. Improvements are smaller after we have 12 months of data, but they still keep improving to a lesser extent. It is much later, after the 21st month, that the performance results start to plateau. Differences between using 21 and 24 months are very small, which suggests that it is the optimal result we can get by using this approach. There are, in fact, cases where the classifier performs even better with 21 months of training data than with 24, especially for very early detection of hoaxes for small values of $t$.

\begin{table*}[tbh]
 \begin{center}
  \begin{tabular}{l c c c c c c c c c c}
  \toprule
  months & 0 & 1' & 2' & 5' & 10' & 15' & 30' & 60' & 120' & 300' \\
  \midrule
  3 & .533 & .560 & .567 & .538 & .518 & .523 & .525 & .550 & .560 & .602 \\
  6 & .608 & .627 & .626 & .642 & .641 & .645 & .652 & .667 & .663 & .676 \\
  9 & .632 & .634 & .644 & .641 & .648 & .665 & .664 & .686 & .692 & .691 \\
  12 & .637 & .666 & .666 & .678 & .687 & .691 & .697 & .709 & .710 & .721 \\
  15 & .645 & .663 & .679 & .690 & .701 & .713 & .721 & .733 & .732 & .730 \\
  18 & .643 & .668 & .683 & .695 & .708 & .714 & .718 & .731 & .735 & .733 \\
  21 & \textbf{.649} & .675 & .689 & .699 & \textbf{.718} & .720 & \textbf{.728} & .737 & .744 & .744 \\
  24 & .647 & \textbf{.677} & \textbf{.696} & \textbf{.707} & .716 & \textbf{.725} & .724 & \textbf{.744} & \textbf{.752} & \textbf{.748} \\
  \bottomrule
  \end{tabular}
  \caption{Performance results for early detection of hoaxes by using different sizes of training data, in months, using the best performing set of features (social+multiw2v).}
  \label{tab:training}
 \end{center}
\end{table*}

For more results on different sizes of the training data using other baseline classifiers, please refer to Appendix \ref{ap:baselines}.

\subsection{Effect of the Data Sampling Strategy}

In the process of generating our dataset of death reports, we made the decision of setting 50 as the minimum number of tweets that a report would need to reach in order to be included in the dataset. This decision was made for scalability issues and for making sure that we have enough data for each report. To determine the impact of setting 50 as the value for this threshold, here we experiment with two other thresholds, 100 and 150. The objective is to see if we get similar results or instead a different threshold would lead to different conclusions.

\begin{figure}[htb]
 \begin{center}
  \includegraphics[width=0.95\textwidth]{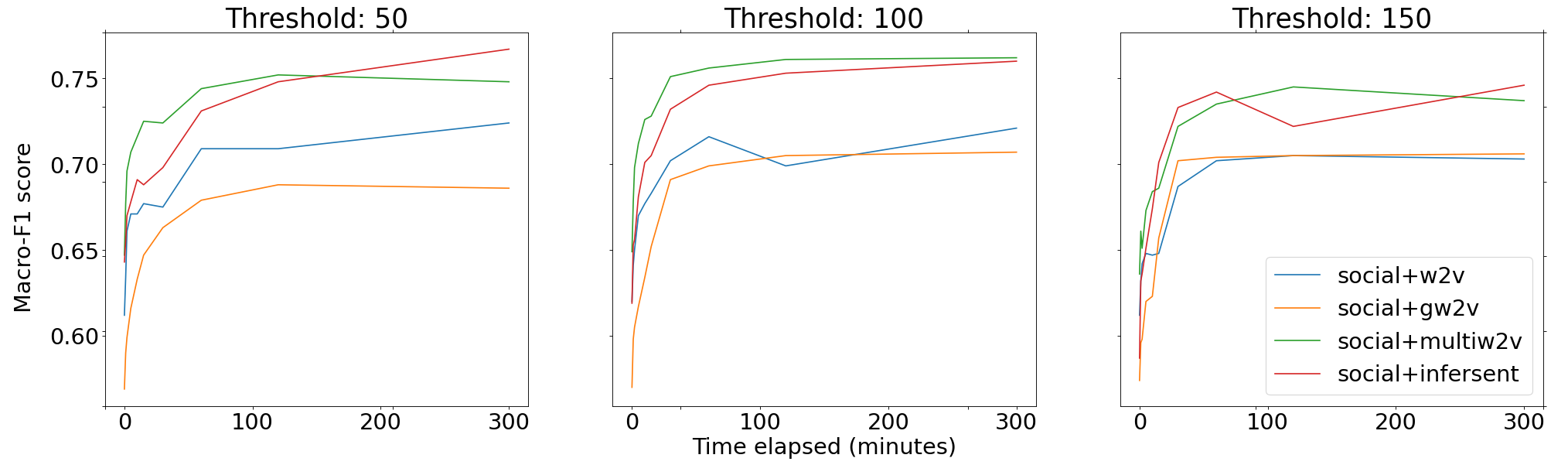}
  \Description{Performance across different sampling thresholds.}
  \caption{Performance across different sampling thresholds.}
  \label{fig:thresholds}
 \end{center}
\end{figure}

Figure \ref{fig:thresholds} shows slight variations as the threshold changes. The use of `social+multiw2v' and `social+infersent' features consistently perform better than the other features. While there are occasions where `social+infersent' can perform best, `social+multiw2v' generally achieves the best performance, particularly in earlier stages of the reports, i.e. in the first 100 minutes. When the threshold is increased to 150, the performance of `social+multiw2v' is slightly below `social+infersent' for the first few minutes only.

\section{Analysis of Features}

\begin{figure}[htb]
 \begin{center}
  \includegraphics[width=0.95\textwidth]{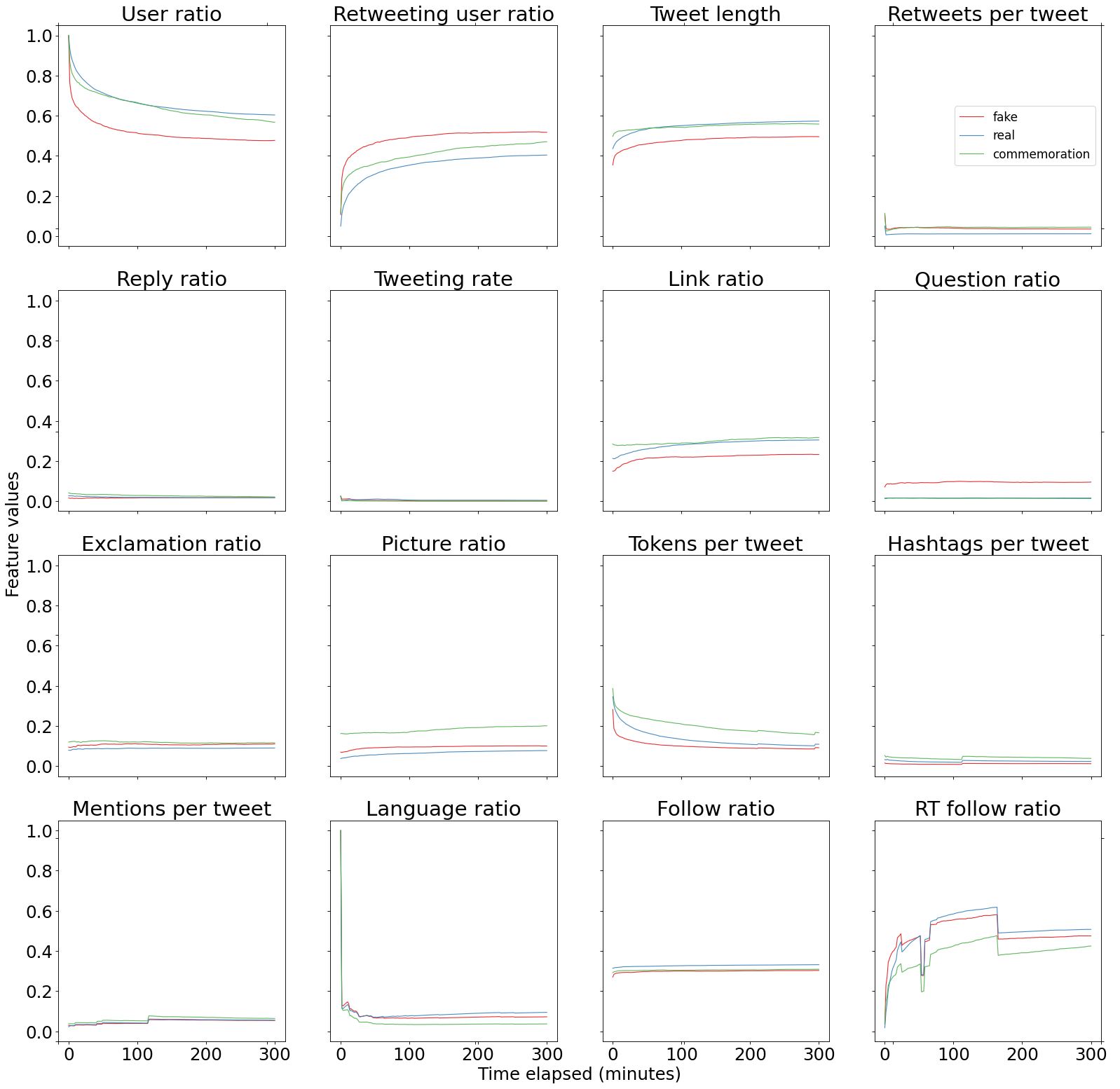}
  \Description{Temporal visualisation of social feature values, comparing across labels: real, commemoration and fake.}
  \caption{Temporal visualisation of social feature values, comparing across labels: real, commemoration and fake.}
  \label{fig:social-features}
 \end{center}
\end{figure}

Figure \ref{fig:social-features} shows the values for the 16 social features in our experiments, plotted as a timeline showing their values over time per category. Feature values are averaged across all of the instances for a particular category. Some figures show a similar increasing/decreasing tendency of values across categories, which is however affected by the normalisation of values we perform. These figures are especially useful to distinguish the values across categories.

We are particularly interested in looking at the cases where fake deaths (red lines) exhibit very different values with respect to real and commemorative deaths (blue and green lines). Some interesting findings we observe from this analysis:
\begin{itemize}
 \item The ratio of distinct users (user ratio) is lower for death hoaxes than it is for the other categories, showing that the number of users who participate in fake stories tends to be lower.
 \item This differs, however, from the the ratio of distinct retweeting users (retweeting user ratio). Here we observe instead that the number of distinct users retweeting death hoaxes tends to be higher than for the other two categories, hence showing that death hoaxes are initiated by a few but retweeted by many.
 \item Tweets associated with death hoaxes tend to be shorter (tweet length) and have fewer words (token ratio), possibly indicating the provision of less context or evidence associated with the report. This ties in with the finding that death hoaxes are less likely to provide links (link ratio), likely owing to the lack of news articles covering those.
 \item Death hoaxes tend to spark more questions (question ratio), which may indicate more skepticism from participating users.
 \item It is the case of commemorations that is more often accompanied with pictures (picture ratio), than for real or fake deaths. This indicates a higher likelihood to commemorate people from the past with memorable pictures.
\end{itemize}

\section{Discussion and Conclusion}

We have introduced a novel approach for semi-automated generation of annotated social media datasets made of celebrity death reports for veracity classification. Different from previous work, our approach does not need to collect true and false stories using different approaches, and consequently enables experimentation in a realistic scenario with a realistic ratio of false stories. Our semi-automated approach consists in leveraging the Wikidata knowledge base, with which we can easily verify if celebrity death reports circulating in social media refer to people who have actually died or are instead made up reports. Following this process, we have produced a dataset comprising 4,007 different death reports, which include over 13 million tweets, and have a ratio of 15\% false stories.

The generation of this dataset has also enabled us to run experiments for early hoax detection from social media, which we have experimented for very early detection within minutes of the first report. Taking advantage of the large-scale of our dataset, we have experimented using class-specific representations of word embeddings. This approach has proven to clearly outperform the use of a single model of word embeddings for the entire dataset. Our approach achieves competitive results for detection of hoaxes within the first 2 to 10 minutes, with F1 scores close to 72\% within 10 minutes. This method based on class-specific embeddings for early detection of hoaxes outperforms state-of-the-art methods for word representation, including word embeddings (w2v) and sentence embeddings (infersent). Different from the latter, our proposed method, multiw2v, is able to leverage the class labels from a large-scale, annotated dataset, to learn different meanings of words across categories, e.g. RIP meaning `Rest in Peace' or `Really Inspiring Person', depending on the type of report. While we have tested the multiw2v word representation approach for the hoax detection task, it is directly applicable to any other classification task where a large-scale annotated dataset is available for building the multiw2v model, for instance thanks to the availability of distantly supervised datasets.

With further experimentation, we have observed that the use of sliding windows, where the most recent tweets are considered for the classification task, is not helpful in this task, and instead using the entire timeline of tweets is better. We have observed that the larger the sliding window, the better is our system's performance, with optimal results for a sliding window covering 100\% of the stream, i.e. the equivalent of not having a sliding window. Finally, we have explored the effect of having different sizes of training sets, showing that performance results start to plateau after 21 months of training data and more training data may not necessarily lead to improved results.

The dataset and the word embedding models developed in this work are publicly available,\footnote{\url{https://figshare.com/articles/Twitter_Death_Hoaxes_dataset/5688811}} enabling further research in this much needed research area using a benchmark dataset.

Our data collection, annotation and experimentation is limited to a specific kind of hoax triggered by death reports. Our motivation to focus on this kind of reports was both their prominence in social media and the need to tackle them, as well as the possibility of modelling the problem by leveraging names of notable people from Wikipedia. Hence this led to the development of a novel method to develop a large-scale dataset to tackle hoaxes, while being restricted to this specific kind of hoaxes. This has the limitation of its direct applicability to broader types of hoaxes, which this work does not cover and is left for future work. This, in turn, should be taken into account when interpreting the findings of this work, whose generalisation to other kinds of hoaxes needs further investigation.

Our plans for future work include experimentation with other events that can be linked to Wikidata or other knowledge bases, beyond death reports, such as resignation of public figures, numbers of casualties reported for emergency events, or other factual claims.

\begin{acks}
 This research utilised Queen Mary's Apocrita HPC facility, supported by QMUL Research-IT. http://doi.org/10.5281/zenodo.438045
\end{acks}

\bibliographystyle{ACM-Reference-Format}
\bibliography{rip}


\begin{thebibliography}{69}


\ifx \showCODEN    \undefined \def \showCODEN     #1{\unskip}     \fi
\ifx \showDOI      \undefined \def \showDOI       #1{#1}\fi
\ifx \showISBNx    \undefined \def \showISBNx     #1{\unskip}     \fi
\ifx \showISBNxiii \undefined \def \showISBNxiii  #1{\unskip}     \fi
\ifx \showISSN     \undefined \def \showISSN      #1{\unskip}     \fi
\ifx \showLCCN     \undefined \def \showLCCN      #1{\unskip}     \fi
\ifx \shownote     \undefined \def \shownote      #1{#1}          \fi
\ifx \showarticletitle \undefined \def \showarticletitle #1{#1}   \fi
\ifx \showURL      \undefined \def \showURL       {\relax}        \fi
\providecommand\bibfield[2]{#2}
\providecommand\bibinfo[2]{#2}
\providecommand\natexlab[1]{#1}
\providecommand\showeprint[2][]{arXiv:#2}

\bibitem[\protect\citeauthoryear{Alkhodair, Ding, Fung, and Liu}{Alkhodair
  et~al\mbox{.}}{2020}]%
        {alkhodair2020detecting}
\bibfield{author}{\bibinfo{person}{Sarah~A Alkhodair},
  \bibinfo{person}{Steven~HH Ding}, \bibinfo{person}{Benjamin~CM Fung}, {and}
  \bibinfo{person}{Junqiang Liu}.} \bibinfo{year}{2020}\natexlab{}.
\newblock \showarticletitle{Detecting breaking news rumors of emerging topics
  in social media}.
\newblock \bibinfo{journal}{\emph{Information Processing \& Management}}
  \bibinfo{volume}{57}, \bibinfo{number}{2} (\bibinfo{year}{2020}),
  \bibinfo{pages}{102018}.
\newblock


\bibitem[\protect\citeauthoryear{Allcott and Gentzkow}{Allcott and
  Gentzkow}{2017}]%
        {allcott2017social}
\bibfield{author}{\bibinfo{person}{Hunt Allcott} {and} \bibinfo{person}{Matthew
  Gentzkow}.} \bibinfo{year}{2017}\natexlab{}.
\newblock \bibinfo{booktitle}{\emph{Social media and fake news in the 2016
  election}}.
\newblock \bibinfo{type}{{T}echnical {R}eport}. \bibinfo{institution}{National
  Bureau of Economic Research}.
\newblock


\bibitem[\protect\citeauthoryear{Bengio, Ducharme, Vincent, and Jauvin}{Bengio
  et~al\mbox{.}}{2003}]%
        {bengio2003neural}
\bibfield{author}{\bibinfo{person}{Yoshua Bengio}, \bibinfo{person}{R{\'e}jean
  Ducharme}, \bibinfo{person}{Pascal Vincent}, {and} \bibinfo{person}{Christian
  Jauvin}.} \bibinfo{year}{2003}\natexlab{}.
\newblock \showarticletitle{A neural probabilistic language model}.
\newblock \bibinfo{journal}{\emph{Journal of machine learning research}}
  \bibinfo{volume}{3}, \bibinfo{number}{Feb} (\bibinfo{year}{2003}),
  \bibinfo{pages}{1137--1155}.
\newblock


\bibitem[\protect\citeauthoryear{Bowman, Angeli, Potts, and Manning}{Bowman
  et~al\mbox{.}}{2015}]%
        {bowman2015large}
\bibfield{author}{\bibinfo{person}{Samuel~R Bowman}, \bibinfo{person}{Gabor
  Angeli}, \bibinfo{person}{Christopher Potts}, {and}
  \bibinfo{person}{Christopher~D Manning}.} \bibinfo{year}{2015}\natexlab{}.
\newblock \showarticletitle{A large annotated corpus for learning natural
  language inference}. In \bibinfo{booktitle}{\emph{Conference on Empirical
  Methods in Natural Language Processing, EMNLP 2015}}. Association for
  Computational Linguistics (ACL).
\newblock


\bibitem[\protect\citeauthoryear{Brown, Desouza, Mercer, Pietra, and Lai}{Brown
  et~al\mbox{.}}{1992}]%
        {brown1992class}
\bibfield{author}{\bibinfo{person}{Peter~F Brown}, \bibinfo{person}{Peter~V
  Desouza}, \bibinfo{person}{Robert~L Mercer}, \bibinfo{person}{Vincent J~Della
  Pietra}, {and} \bibinfo{person}{Jenifer~C Lai}.}
  \bibinfo{year}{1992}\natexlab{}.
\newblock \showarticletitle{Class-based n-gram models of natural language}.
\newblock \bibinfo{journal}{\emph{Computational linguistics}}
  \bibinfo{volume}{18}, \bibinfo{number}{4} (\bibinfo{year}{1992}),
  \bibinfo{pages}{467--479}.
\newblock


\bibitem[\protect\citeauthoryear{Bruns, Highfield, and Lind}{Bruns
  et~al\mbox{.}}{2012}]%
        {bruns2012blogs}
\bibfield{author}{\bibinfo{person}{Axel Bruns}, \bibinfo{person}{Tim
  Highfield}, {and} \bibinfo{person}{Rebecca~Ann Lind}.}
  \bibinfo{year}{2012}\natexlab{}.
\newblock \showarticletitle{Blogs, Twitter, and breaking news: The produsage of
  citizen journalism}.
\newblock \bibinfo{journal}{\emph{Produsing theory in a digital world: The
  intersection of audiences and production in contemporary theory}}
  \bibinfo{volume}{80}, \bibinfo{number}{2012} (\bibinfo{year}{2012}),
  \bibinfo{pages}{15--32}.
\newblock


\bibitem[\protect\citeauthoryear{Castillo, Mendoza, and Poblete}{Castillo
  et~al\mbox{.}}{2011}]%
        {castillo2011information}
\bibfield{author}{\bibinfo{person}{Carlos Castillo}, \bibinfo{person}{Marcelo
  Mendoza}, {and} \bibinfo{person}{Barbara Poblete}.}
  \bibinfo{year}{2011}\natexlab{}.
\newblock \showarticletitle{Information credibility on twitter}. In
  \bibinfo{booktitle}{\emph{Proceedings of the 20th international conference on
  World wide web}}. ACM, \bibinfo{pages}{675--684}.
\newblock


\bibitem[\protect\citeauthoryear{Cohen}{Cohen}{1960}]%
        {cohen1960coefficient}
\bibfield{author}{\bibinfo{person}{Jacob Cohen}.}
  \bibinfo{year}{1960}\natexlab{}.
\newblock \showarticletitle{A coefficient of agreement for nominal scales}.
\newblock \bibinfo{journal}{\emph{Educational and psychological measurement}}
  \bibinfo{volume}{20}, \bibinfo{number}{1} (\bibinfo{year}{1960}),
  \bibinfo{pages}{37--46}.
\newblock


\bibitem[\protect\citeauthoryear{Conneau, Kiela, Schwenk, Barrault, and
  Bordes}{Conneau et~al\mbox{.}}{2017}]%
        {conneau2017supervised}
\bibfield{author}{\bibinfo{person}{Alexis Conneau}, \bibinfo{person}{Douwe
  Kiela}, \bibinfo{person}{Holger Schwenk}, \bibinfo{person}{Lo{\"\i}c
  Barrault}, {and} \bibinfo{person}{Antoine Bordes}.}
  \bibinfo{year}{2017}\natexlab{}.
\newblock \showarticletitle{Supervised Learning of Universal Sentence
  Representations from Natural Language Inference Data}. In
  \bibinfo{booktitle}{\emph{Proceedings of the 2017 Conference on Empirical
  Methods in Natural Language Processing}}. \bibinfo{pages}{670--680}.
\newblock


\bibitem[\protect\citeauthoryear{Datar, Gionis, Indyk, and Motwani}{Datar
  et~al\mbox{.}}{2002}]%
        {datar2002maintaining}
\bibfield{author}{\bibinfo{person}{Mayur Datar}, \bibinfo{person}{Aristides
  Gionis}, \bibinfo{person}{Piotr Indyk}, {and} \bibinfo{person}{Rajeev
  Motwani}.} \bibinfo{year}{2002}\natexlab{}.
\newblock \showarticletitle{Maintaining stream statistics over sliding
  windows}.
\newblock \bibinfo{journal}{\emph{SIAM journal on computing}}
  \bibinfo{volume}{31}, \bibinfo{number}{6} (\bibinfo{year}{2002}),
  \bibinfo{pages}{1794--1813}.
\newblock


\bibitem[\protect\citeauthoryear{Derczynski, Bontcheva, Liakata, Procter, Wong
  Sak~Hoi, and Zubiaga}{Derczynski et~al\mbox{.}}{2017}]%
        {derczynski2017semeval}
\bibfield{author}{\bibinfo{person}{Leon Derczynski}, \bibinfo{person}{Kalina
  Bontcheva}, \bibinfo{person}{Maria Liakata}, \bibinfo{person}{Rob Procter},
  \bibinfo{person}{Geraldine Wong Sak~Hoi}, {and} \bibinfo{person}{Arkaitz
  Zubiaga}.} \bibinfo{year}{2017}\natexlab{}.
\newblock \showarticletitle{{SemEval-2017 Task 8: RumourEval: Determining
  rumour veracity and support for rumours}}. In
  \bibinfo{booktitle}{\emph{Proceedings of SemEval}}. ACL,
  \bibinfo{pages}{69--76}.
\newblock


\bibitem[\protect\citeauthoryear{Diakopoulos, De~Choudhury, and
  Naaman}{Diakopoulos et~al\mbox{.}}{2012}]%
        {diakopoulos2012finding}
\bibfield{author}{\bibinfo{person}{Nicholas Diakopoulos},
  \bibinfo{person}{Munmun De~Choudhury}, {and} \bibinfo{person}{Mor Naaman}.}
  \bibinfo{year}{2012}\natexlab{}.
\newblock \showarticletitle{Finding and assessing social media information
  sources in the context of journalism}. In
  \bibinfo{booktitle}{\emph{Proceedings of CHI}}. ACM,
  \bibinfo{pages}{2451--2460}.
\newblock


\bibitem[\protect\citeauthoryear{Dungs, Aker, Fuhr, and Bontcheva}{Dungs
  et~al\mbox{.}}{2018}]%
        {dungs2018can}
\bibfield{author}{\bibinfo{person}{Sebastian Dungs}, \bibinfo{person}{Ahmet
  Aker}, \bibinfo{person}{Norbert Fuhr}, {and} \bibinfo{person}{Kalina
  Bontcheva}.} \bibinfo{year}{2018}\natexlab{}.
\newblock \showarticletitle{Can rumour stance alone predict veracity?}. In
  \bibinfo{booktitle}{\emph{Proceedings of the 27th International Conference on
  Computational Linguistics}}. \bibinfo{pages}{3360--3370}.
\newblock


\bibitem[\protect\citeauthoryear{Gerbaudo}{Gerbaudo}{2018}]%
        {gerbaudo2018tweets}
\bibfield{author}{\bibinfo{person}{Paolo Gerbaudo}.}
  \bibinfo{year}{2018}\natexlab{}.
\newblock \bibinfo{booktitle}{\emph{Tweets and the streets: Social media and
  contemporary activism}}.
\newblock \bibinfo{publisher}{Pluto Press}.
\newblock


\bibitem[\protect\citeauthoryear{Gereme and Zhu}{Gereme and Zhu}{2019}]%
        {gereme2019early}
\bibfield{author}{\bibinfo{person}{Fantahun~Bogale Gereme} {and}
  \bibinfo{person}{William Zhu}.} \bibinfo{year}{2019}\natexlab{}.
\newblock \showarticletitle{Early Detection of Fake News" Before It Flies
  High"}. In \bibinfo{booktitle}{\emph{Proceedings of the 2nd International
  Conference on Big Data Technologies}}. \bibinfo{pages}{142--148}.
\newblock


\bibitem[\protect\citeauthoryear{Golbeck, Mauriello, Auxier, Bhanushali, Bonk,
  Bouzaghrane, Buntain, Chanduka, Cheakalos, Everett, et~al\mbox{.}}{Golbeck
  et~al\mbox{.}}{2018}]%
        {golbeck2018fake}
\bibfield{author}{\bibinfo{person}{Jennifer Golbeck}, \bibinfo{person}{Matthew
  Mauriello}, \bibinfo{person}{Brooke Auxier}, \bibinfo{person}{Keval~H
  Bhanushali}, \bibinfo{person}{Christopher Bonk},
  \bibinfo{person}{Mohamed~Amine Bouzaghrane}, \bibinfo{person}{Cody Buntain},
  \bibinfo{person}{Riya Chanduka}, \bibinfo{person}{Paul Cheakalos},
  \bibinfo{person}{Jennine~B Everett}, {et~al\mbox{.}}}
  \bibinfo{year}{2018}\natexlab{}.
\newblock \showarticletitle{Fake news vs satire: A dataset and analysis}. In
  \bibinfo{booktitle}{\emph{Proceedings of the 10th ACM Conference on Web
  Science}}. \bibinfo{pages}{17--21}.
\newblock


\bibitem[\protect\citeauthoryear{Gorrell, Bontcheva, Derczynski, Kochkina,
  Liakata, and Zubiaga}{Gorrell et~al\mbox{.}}{2019}]%
        {gorrell2018rumoureval}
\bibfield{author}{\bibinfo{person}{Genevieve Gorrell}, \bibinfo{person}{Kalina
  Bontcheva}, \bibinfo{person}{Leon Derczynski}, \bibinfo{person}{Elena
  Kochkina}, \bibinfo{person}{Maria Liakata}, {and} \bibinfo{person}{Arkaitz
  Zubiaga}.} \bibinfo{year}{2019}\natexlab{}.
\newblock \showarticletitle{{RumourEval 2019: Determining Rumour Veracity and
  Support for Rumours}}.
\newblock \bibinfo{journal}{\emph{Proceedings of SemEval}}
  (\bibinfo{year}{2019}), \bibinfo{pages}{845--854}.
\newblock


\bibitem[\protect\citeauthoryear{Gottfried and Shearer}{Gottfried and
  Shearer}{2016}]%
        {gottfried2016news}
\bibfield{author}{\bibinfo{person}{Jeffrey Gottfried} {and}
  \bibinfo{person}{Elisa Shearer}.} \bibinfo{year}{2016}\natexlab{}.
\newblock \bibinfo{booktitle}{\emph{News Use Across Social Media Platforms
  2016}}.
\newblock \bibinfo{type}{{T}echnical {R}eport}. \bibinfo{institution}{Pew
  Research Center}.
\newblock


\bibitem[\protect\citeauthoryear{Gupta, Kumaraguru, Castillo, and Meier}{Gupta
  et~al\mbox{.}}{2014}]%
        {gupta2014tweetcred}
\bibfield{author}{\bibinfo{person}{Aditi Gupta}, \bibinfo{person}{Ponnurangam
  Kumaraguru}, \bibinfo{person}{Carlos Castillo}, {and}
  \bibinfo{person}{Patrick Meier}.} \bibinfo{year}{2014}\natexlab{}.
\newblock \showarticletitle{Tweetcred: Real-time credibility assessment of
  content on twitter}. In \bibinfo{booktitle}{\emph{International Conference on
  Social Informatics}}. Springer, \bibinfo{pages}{228--243}.
\newblock


\bibitem[\protect\citeauthoryear{Hermida}{Hermida}{2012}]%
        {hermida2012tweets}
\bibfield{author}{\bibinfo{person}{Alfred Hermida}.}
  \bibinfo{year}{2012}\natexlab{}.
\newblock \showarticletitle{Tweets and truth: Journalism as a discipline of
  collaborative verification}.
\newblock \bibinfo{journal}{\emph{Journalism Practice}} \bibinfo{volume}{6},
  \bibinfo{number}{5-6} (\bibinfo{year}{2012}), \bibinfo{pages}{659--668}.
\newblock


\bibitem[\protect\citeauthoryear{Jin, Cao, Zhang, and Luo}{Jin
  et~al\mbox{.}}{2016}]%
        {jin2016news}
\bibfield{author}{\bibinfo{person}{Zhiwei Jin}, \bibinfo{person}{Juan Cao},
  \bibinfo{person}{Yongdong Zhang}, {and} \bibinfo{person}{Jiebo Luo}.}
  \bibinfo{year}{2016}\natexlab{}.
\newblock \showarticletitle{News Verification by Exploiting Conflicting Social
  Viewpoints in Microblogs.}. In \bibinfo{booktitle}{\emph{AAAI}}.
  \bibinfo{pages}{2972--2978}.
\newblock


\bibitem[\protect\citeauthoryear{Jindal, Vatsa, and Singh}{Jindal
  et~al\mbox{.}}{2019}]%
        {jindal2019newsbag}
\bibfield{author}{\bibinfo{person}{Sarthak Jindal}, \bibinfo{person}{Mayank
  Vatsa}, {and} \bibinfo{person}{Richa Singh}.}
  \bibinfo{year}{2019}\natexlab{}.
\newblock \showarticletitle{Newsbag: a benchmark dataset for fake news
  detection}.
\newblock  (\bibinfo{year}{2019}).
\newblock


\bibitem[\protect\citeauthoryear{Koo, Carreras~P{\'e}rez, and Collins}{Koo
  et~al\mbox{.}}{2008}]%
        {koo2008simple}
\bibfield{author}{\bibinfo{person}{Terry Koo}, \bibinfo{person}{Xavier
  Carreras~P{\'e}rez}, {and} \bibinfo{person}{Michael Collins}.}
  \bibinfo{year}{2008}\natexlab{}.
\newblock \showarticletitle{Simple semi-supervised dependency parsing}. In
  \bibinfo{booktitle}{\emph{Proceedings of ACL}}. \bibinfo{pages}{595--603}.
\newblock


\bibitem[\protect\citeauthoryear{Kumar, West, and Leskovec}{Kumar
  et~al\mbox{.}}{2016}]%
        {kumar2016disinformation}
\bibfield{author}{\bibinfo{person}{Srijan Kumar}, \bibinfo{person}{Robert
  West}, {and} \bibinfo{person}{Jure Leskovec}.}
  \bibinfo{year}{2016}\natexlab{}.
\newblock \showarticletitle{Disinformation on the web: Impact, characteristics,
  and detection of wikipedia hoaxes}. In \bibinfo{booktitle}{\emph{Proceedings
  of WWW}}. \bibinfo{pages}{591--602}.
\newblock


\bibitem[\protect\citeauthoryear{Kwak, Lee, Park, and Moon}{Kwak
  et~al\mbox{.}}{2010}]%
        {kwak2010twitter}
\bibfield{author}{\bibinfo{person}{Haewoon Kwak}, \bibinfo{person}{Changhyun
  Lee}, \bibinfo{person}{Hosung Park}, {and} \bibinfo{person}{Sue Moon}.}
  \bibinfo{year}{2010}\natexlab{}.
\newblock \showarticletitle{What is Twitter, a social network or a news
  media?}. In \bibinfo{booktitle}{\emph{Proceedings of WWW}}. ACM,
  \bibinfo{pages}{591--600}.
\newblock


\bibitem[\protect\citeauthoryear{Kwon, Cha, and Jung}{Kwon
  et~al\mbox{.}}{2017}]%
        {kwon2017rumor}
\bibfield{author}{\bibinfo{person}{Sejeong Kwon}, \bibinfo{person}{Meeyoung
  Cha}, {and} \bibinfo{person}{Kyomin Jung}.} \bibinfo{year}{2017}\natexlab{}.
\newblock \showarticletitle{Rumor detection over varying time windows}.
\newblock \bibinfo{journal}{\emph{PloS one}} \bibinfo{volume}{12},
  \bibinfo{number}{1} (\bibinfo{year}{2017}), \bibinfo{pages}{e0168344}.
\newblock


\bibitem[\protect\citeauthoryear{Liu, Nourbakhsh, Li, Fang, and Shah}{Liu
  et~al\mbox{.}}{2015}]%
        {liu2015real}
\bibfield{author}{\bibinfo{person}{Xiaomo Liu}, \bibinfo{person}{Armineh
  Nourbakhsh}, \bibinfo{person}{Quanzhi Li}, \bibinfo{person}{Rui Fang}, {and}
  \bibinfo{person}{Sameena Shah}.} \bibinfo{year}{2015}\natexlab{}.
\newblock \showarticletitle{Real-time rumor debunking on twitter}. In
  \bibinfo{booktitle}{\emph{Proceedings of CIKM}}. ACM,
  \bibinfo{pages}{1867--1870}.
\newblock


\bibitem[\protect\citeauthoryear{Liu and Wu}{Liu and Wu}{2018}]%
        {liu2018early}
\bibfield{author}{\bibinfo{person}{Yang Liu} {and}
  \bibinfo{person}{Yi-Fang~Brook Wu}.} \bibinfo{year}{2018}\natexlab{}.
\newblock \showarticletitle{Early detection of fake news on social media
  through propagation path classification with recurrent and convolutional
  networks}. In \bibinfo{booktitle}{\emph{Thirty-Second AAAI Conference on
  Artificial Intelligence}}.
\newblock


\bibitem[\protect\citeauthoryear{Ma, Gao, Mitra, Kwon, Jansen, Wong, and
  Cha}{Ma et~al\mbox{.}}{2016}]%
        {ma2016detecting}
\bibfield{author}{\bibinfo{person}{Jing Ma}, \bibinfo{person}{Wei Gao},
  \bibinfo{person}{Prasenjit Mitra}, \bibinfo{person}{Sejeong Kwon},
  \bibinfo{person}{Bernard~J Jansen}, \bibinfo{person}{Kam-Fai Wong}, {and}
  \bibinfo{person}{Meeyoung Cha}.} \bibinfo{year}{2016}\natexlab{}.
\newblock \showarticletitle{Detecting Rumors from Microblogs with Recurrent
  Neural Networks.}. In \bibinfo{booktitle}{\emph{IJCAI}}.
  \bibinfo{pages}{3818--3824}.
\newblock


\bibitem[\protect\citeauthoryear{MacDougall}{MacDougall}{1958}]%
        {macdougall1958hoaxes}
\bibfield{author}{\bibinfo{person}{Curtis~Daniel MacDougall}.}
  \bibinfo{year}{1958}\natexlab{}.
\newblock \bibinfo{booktitle}{\emph{Hoaxes}}. Vol.~\bibinfo{volume}{465}.
\newblock \bibinfo{publisher}{Dover Pubns}.
\newblock


\bibitem[\protect\citeauthoryear{Menczer}{Menczer}{2016}]%
        {menczer2016spread}
\bibfield{author}{\bibinfo{person}{Filippo Menczer}.}
  \bibinfo{year}{2016}\natexlab{}.
\newblock \showarticletitle{The spread of misinformation in social media}. In
  \bibinfo{booktitle}{\emph{Proceedings of WWW}}. \bibinfo{pages}{717--717}.
\newblock


\bibitem[\protect\citeauthoryear{Mikolov, Sutskever, Chen, Corrado, and
  Dean}{Mikolov et~al\mbox{.}}{2013}]%
        {mikolov2013distributed}
\bibfield{author}{\bibinfo{person}{Tomas Mikolov}, \bibinfo{person}{Ilya
  Sutskever}, \bibinfo{person}{Kai Chen}, \bibinfo{person}{Greg~S Corrado},
  {and} \bibinfo{person}{Jeff Dean}.} \bibinfo{year}{2013}\natexlab{}.
\newblock \showarticletitle{Distributed representations of words and phrases
  and their compositionality}. In \bibinfo{booktitle}{\emph{Advances in neural
  information processing systems}}. \bibinfo{pages}{3111--3119}.
\newblock


\bibitem[\protect\citeauthoryear{Mitra and Gilbert}{Mitra and Gilbert}{2015}]%
        {mitra2015credbank}
\bibfield{author}{\bibinfo{person}{Tanushree Mitra} {and} \bibinfo{person}{Eric
  Gilbert}.} \bibinfo{year}{2015}\natexlab{}.
\newblock \showarticletitle{Credbank: A large-scale social media corpus with
  associated credibility annotations}. In \bibinfo{booktitle}{\emph{Ninth
  International AAAI Conference on Web and Social Media}}.
\newblock


\bibitem[\protect\citeauthoryear{Moreno and Bressan}{Moreno and
  Bressan}{2019}]%
        {moreno2019factck}
\bibfield{author}{\bibinfo{person}{Jo{\~a}o Moreno} {and}
  \bibinfo{person}{Gra{\c{c}}a Bressan}.} \bibinfo{year}{2019}\natexlab{}.
\newblock \showarticletitle{FACTCK. BR: a new dataset to study fake news}. In
  \bibinfo{booktitle}{\emph{Proceedings of the 25th Brazillian Symposium on
  Multimedia and the Web}}. \bibinfo{pages}{525--527}.
\newblock


\bibitem[\protect\citeauthoryear{Morris, Counts, Roseway, Hoff, and
  Schwarz}{Morris et~al\mbox{.}}{2012}]%
        {morris2012tweeting}
\bibfield{author}{\bibinfo{person}{Meredith~Ringel Morris},
  \bibinfo{person}{Scott Counts}, \bibinfo{person}{Asta Roseway},
  \bibinfo{person}{Aaron Hoff}, {and} \bibinfo{person}{Julia Schwarz}.}
  \bibinfo{year}{2012}\natexlab{}.
\newblock \showarticletitle{Tweeting is believing?: understanding microblog
  credibility perceptions}. In \bibinfo{booktitle}{\emph{Proceedings of CSCW}}.
  ACM, \bibinfo{pages}{441--450}.
\newblock


\bibitem[\protect\citeauthoryear{Nakamura, Levy, and Wang}{Nakamura
  et~al\mbox{.}}{2019}]%
        {nakamura2019r}
\bibfield{author}{\bibinfo{person}{Kai Nakamura}, \bibinfo{person}{Sharon
  Levy}, {and} \bibinfo{person}{William~Yang Wang}.}
  \bibinfo{year}{2019}\natexlab{}.
\newblock \showarticletitle{r/Fakeddit: A New Multimodal Benchmark Dataset for
  Fine-grained Fake News Detection}.
\newblock \bibinfo{journal}{\emph{arXiv preprint arXiv:1911.03854}}
  (\bibinfo{year}{2019}).
\newblock


\bibitem[\protect\citeauthoryear{Nares}{Nares}{1822}]%
        {nares1822glossary}
\bibfield{author}{\bibinfo{person}{Robert Nares}.}
  \bibinfo{year}{1822}\natexlab{}.
\newblock \bibinfo{booktitle}{\emph{A Glossary: Or, Collection of Words,
  Phrases, Names, and Allusions to Customs, Proverbs, \&c., which Have Been
  Thought to Require Illustration, in the Works of English Authors,
  Particularly Shakespeare, and His Contemporaries...}}
\newblock \bibinfo{publisher}{R. Triphook}.
\newblock


\bibitem[\protect\citeauthoryear{N{\o}rregaard, Horne, and
  Adal{\i}}{N{\o}rregaard et~al\mbox{.}}{2019}]%
        {norregaard2019nela}
\bibfield{author}{\bibinfo{person}{Jeppe N{\o}rregaard},
  \bibinfo{person}{Benjamin~D Horne}, {and} \bibinfo{person}{Sibel Adal{\i}}.}
  \bibinfo{year}{2019}\natexlab{}.
\newblock \showarticletitle{NELA-GT-2018: A large multi-labelled news dataset
  for the study of misinformation in news articles}. In
  \bibinfo{booktitle}{\emph{Proceedings of the International AAAI Conference on
  Web and Social Media}}, Vol.~\bibinfo{volume}{13}. \bibinfo{pages}{630--638}.
\newblock


\bibitem[\protect\citeauthoryear{Pedregosa, Varoquaux, Gramfort, Michel,
  Thirion, Grisel, Blondel, Prettenhofer, Weiss, Dubourg, Vanderplas, Passos,
  Cournapeau, Brucher, Perrot, and Duchesnay}{Pedregosa et~al\mbox{.}}{2011}]%
        {pedregosa2011scikit}
\bibfield{author}{\bibinfo{person}{F. Pedregosa}, \bibinfo{person}{G.
  Varoquaux}, \bibinfo{person}{A. Gramfort}, \bibinfo{person}{V. Michel},
  \bibinfo{person}{B. Thirion}, \bibinfo{person}{O. Grisel},
  \bibinfo{person}{M. Blondel}, \bibinfo{person}{P. Prettenhofer},
  \bibinfo{person}{R. Weiss}, \bibinfo{person}{V. Dubourg}, \bibinfo{person}{J.
  Vanderplas}, \bibinfo{person}{A. Passos}, \bibinfo{person}{D. Cournapeau},
  \bibinfo{person}{M. Brucher}, \bibinfo{person}{M. Perrot}, {and}
  \bibinfo{person}{E. Duchesnay}.} \bibinfo{year}{2011}\natexlab{}.
\newblock \showarticletitle{Scikit-learn: Machine Learning in {P}ython}.
\newblock \bibinfo{journal}{\emph{Journal of Machine Learning Research}}
  \bibinfo{volume}{12} (\bibinfo{year}{2011}), \bibinfo{pages}{2825--2830}.
\newblock


\bibitem[\protect\citeauthoryear{Qazvinian, Rosengren, Radev, and
  Mei}{Qazvinian et~al\mbox{.}}{2011}]%
        {qazvinian2011rumor}
\bibfield{author}{\bibinfo{person}{Vahed Qazvinian}, \bibinfo{person}{Emily
  Rosengren}, \bibinfo{person}{Dragomir~R Radev}, {and}
  \bibinfo{person}{Qiaozhu Mei}.} \bibinfo{year}{2011}\natexlab{}.
\newblock \showarticletitle{Rumor has it: Identifying misinformation in
  microblogs}. In \bibinfo{booktitle}{\emph{Proceedings of EMNLP}}.
  \bibinfo{pages}{1589--1599}.
\newblock


\bibitem[\protect\citeauthoryear{Ratnaparkhi}{Ratnaparkhi}{1997}]%
        {ratnaparkhi1997simple}
\bibfield{author}{\bibinfo{person}{Adwait Ratnaparkhi}.}
  \bibinfo{year}{1997}\natexlab{}.
\newblock \showarticletitle{A simple introduction to maximum entropy models for
  natural language processing}.
\newblock \bibinfo{journal}{\emph{IRCS Tech. Reports Series}}
  (\bibinfo{year}{1997}), \bibinfo{pages}{81}.
\newblock


\bibitem[\protect\citeauthoryear{Reis, Correia, Murai, Veloso, Benevenuto, and
  Cambria}{Reis et~al\mbox{.}}{2019}]%
        {reis2019supervised}
\bibfield{author}{\bibinfo{person}{Julio~CS Reis}, \bibinfo{person}{Andr{\'e}
  Correia}, \bibinfo{person}{Fabr{\'\i}cio Murai}, \bibinfo{person}{Adriano
  Veloso}, \bibinfo{person}{Fabr{\'\i}cio Benevenuto}, {and}
  \bibinfo{person}{Erik Cambria}.} \bibinfo{year}{2019}\natexlab{}.
\newblock \showarticletitle{Supervised Learning for Fake News Detection}.
\newblock \bibinfo{journal}{\emph{IEEE Intelligent Systems}}
  \bibinfo{volume}{34}, \bibinfo{number}{2} (\bibinfo{year}{2019}),
  \bibinfo{pages}{76--81}.
\newblock


\bibitem[\protect\citeauthoryear{Salem, Al~Feel, Elbassuoni, Jaber, and
  Farah}{Salem et~al\mbox{.}}{2019}]%
        {salem2019fa}
\bibfield{author}{\bibinfo{person}{Fatima K~Abu Salem}, \bibinfo{person}{Roaa
  Al~Feel}, \bibinfo{person}{Shady Elbassuoni}, \bibinfo{person}{Mohamad
  Jaber}, {and} \bibinfo{person}{May Farah}.} \bibinfo{year}{2019}\natexlab{}.
\newblock \showarticletitle{{FA-KES: a fake news dataset around the Syrian
  war}}. In \bibinfo{booktitle}{\emph{Proceedings of the International AAAI
  Conference on Web and Social Media}}, Vol.~\bibinfo{volume}{13}.
  \bibinfo{pages}{573--582}.
\newblock


\bibitem[\protect\citeauthoryear{Sampson, Morstatter, Wu, and Liu}{Sampson
  et~al\mbox{.}}{2016}]%
        {sampson2016leveraging}
\bibfield{author}{\bibinfo{person}{Justin Sampson}, \bibinfo{person}{Fred
  Morstatter}, \bibinfo{person}{Liang Wu}, {and} \bibinfo{person}{Huan Liu}.}
  \bibinfo{year}{2016}\natexlab{}.
\newblock \showarticletitle{Leveraging the implicit structure within social
  media for emergent rumor detection}. In \bibinfo{booktitle}{\emph{Proceedings
  of CIKM}}. ACM, \bibinfo{pages}{2377--2382}.
\newblock


\bibitem[\protect\citeauthoryear{Sankaranarayanan, Samet, Teitler, Lieberman,
  and Sperling}{Sankaranarayanan et~al\mbox{.}}{2009}]%
        {sankaranarayanan2009twitterstand}
\bibfield{author}{\bibinfo{person}{Jagan Sankaranarayanan},
  \bibinfo{person}{Hanan Samet}, \bibinfo{person}{Benjamin~E Teitler},
  \bibinfo{person}{Michael~D Lieberman}, {and} \bibinfo{person}{Jon Sperling}.}
  \bibinfo{year}{2009}\natexlab{}.
\newblock \showarticletitle{Twitterstand: news in tweets}. In
  \bibinfo{booktitle}{\emph{Proceedings of SIGSPATIAL}}. ACM,
  \bibinfo{pages}{42--51}.
\newblock


\bibitem[\protect\citeauthoryear{Shao, Ciampaglia, Varol, Yang, Flammini, and
  Menczer}{Shao et~al\mbox{.}}{2018}]%
        {shao2018spread}
\bibfield{author}{\bibinfo{person}{Chengcheng Shao},
  \bibinfo{person}{Giovanni~Luca Ciampaglia}, \bibinfo{person}{Onur Varol},
  \bibinfo{person}{Kai-Cheng Yang}, \bibinfo{person}{Alessandro Flammini},
  {and} \bibinfo{person}{Filippo Menczer}.} \bibinfo{year}{2018}\natexlab{}.
\newblock \showarticletitle{The spread of low-credibility content by social
  bots}.
\newblock \bibinfo{journal}{\emph{Nature communications}} \bibinfo{volume}{9},
  \bibinfo{number}{1} (\bibinfo{year}{2018}), \bibinfo{pages}{4787}.
\newblock


\bibitem[\protect\citeauthoryear{Shu, Mahudeswaran, Wang, Lee, and Liu}{Shu
  et~al\mbox{.}}{2018}]%
        {shu2018fakenewsnet}
\bibfield{author}{\bibinfo{person}{Kai Shu}, \bibinfo{person}{Deepak
  Mahudeswaran}, \bibinfo{person}{Suhang Wang}, \bibinfo{person}{Dongwon Lee},
  {and} \bibinfo{person}{Huan Liu}.} \bibinfo{year}{2018}\natexlab{}.
\newblock \showarticletitle{FakeNewsNet: A Data Repository with News Content,
  Social Context and Spatialtemporal Information for Studying Fake News on
  Social Media}.
\newblock \bibinfo{journal}{\emph{arXiv preprint arXiv:1809.01286}}
  (\bibinfo{year}{2018}).
\newblock


\bibitem[\protect\citeauthoryear{Shu, Sliva, Wang, Tang, and Liu}{Shu
  et~al\mbox{.}}{2017}]%
        {shu2017fake}
\bibfield{author}{\bibinfo{person}{Kai Shu}, \bibinfo{person}{Amy Sliva},
  \bibinfo{person}{Suhang Wang}, \bibinfo{person}{Jiliang Tang}, {and}
  \bibinfo{person}{Huan Liu}.} \bibinfo{year}{2017}\natexlab{}.
\newblock \showarticletitle{Fake News Detection on Social Media: A Data Mining
  Perspective}.
\newblock \bibinfo{journal}{\emph{ACM SIGKDD Explorations Newsletter}}
  \bibinfo{volume}{19}, \bibinfo{number}{1} (\bibinfo{year}{2017}),
  \bibinfo{pages}{22--36}.
\newblock


\bibitem[\protect\citeauthoryear{Tacchini, Ballarin, Della~Vedova, Moret, and
  de~Alfaro}{Tacchini et~al\mbox{.}}{2017}]%
        {tacchini2017some}
\bibfield{author}{\bibinfo{person}{Eugenio Tacchini}, \bibinfo{person}{Gabriele
  Ballarin}, \bibinfo{person}{Marco~L Della~Vedova}, \bibinfo{person}{Stefano
  Moret}, {and} \bibinfo{person}{Luca de Alfaro}.}
  \bibinfo{year}{2017}\natexlab{}.
\newblock \showarticletitle{Some Like it Hoax: Automated Fake News Detection in
  Social Networks}. In \bibinfo{booktitle}{\emph{Proceedings of the Second
  Workshop on Data Science for Social Good}}.
\newblock


\bibitem[\protect\citeauthoryear{Takahashi and Igata}{Takahashi and
  Igata}{2012}]%
        {takahashi2012rumor}
\bibfield{author}{\bibinfo{person}{Tetsuro Takahashi} {and}
  \bibinfo{person}{Nobuyuki Igata}.} \bibinfo{year}{2012}\natexlab{}.
\newblock \showarticletitle{Rumor detection on twitter}. In
  \bibinfo{booktitle}{\emph{Proceedings of SCIS}}. IEEE,
  \bibinfo{pages}{452--457}.
\newblock


\bibitem[\protect\citeauthoryear{Thorne, Vlachos, Christodoulopoulos, and
  Mittal}{Thorne et~al\mbox{.}}{2018}]%
        {thorne2018fever}
\bibfield{author}{\bibinfo{person}{James Thorne}, \bibinfo{person}{Andreas
  Vlachos}, \bibinfo{person}{Christos Christodoulopoulos}, {and}
  \bibinfo{person}{Arpit Mittal}.} \bibinfo{year}{2018}\natexlab{}.
\newblock \showarticletitle{FEVER: a Large-scale Dataset for Fact Extraction
  and VERification}. In \bibinfo{booktitle}{\emph{Proceedings of the 2018
  Conference of the North American Chapter of the Association for Computational
  Linguistics: Human Language Technologies, Volume 1 (Long Papers)}}.
  \bibinfo{pages}{809--819}.
\newblock


\bibitem[\protect\citeauthoryear{Tschiatschek, Singla, Gomez~Rodriguez,
  Merchant, and Krause}{Tschiatschek et~al\mbox{.}}{2018}]%
        {tschiatschek2018fake}
\bibfield{author}{\bibinfo{person}{Sebastian Tschiatschek},
  \bibinfo{person}{Adish Singla}, \bibinfo{person}{Manuel Gomez~Rodriguez},
  \bibinfo{person}{Arpit Merchant}, {and} \bibinfo{person}{Andreas Krause}.}
  \bibinfo{year}{2018}\natexlab{}.
\newblock \showarticletitle{Fake news detection in social networks via crowd
  signals}. In \bibinfo{booktitle}{\emph{Companion Proceedings of the The Web
  Conference 2018}}. International World Wide Web Conferences Steering
  Committee, \bibinfo{pages}{517--524}.
\newblock


\bibitem[\protect\citeauthoryear{Turian, Ratinov, and Bengio}{Turian
  et~al\mbox{.}}{2010}]%
        {turian2010word}
\bibfield{author}{\bibinfo{person}{Joseph Turian}, \bibinfo{person}{Lev
  Ratinov}, {and} \bibinfo{person}{Yoshua Bengio}.}
  \bibinfo{year}{2010}\natexlab{}.
\newblock \showarticletitle{Word representations: a simple and general method
  for semi-supervised learning}. In \bibinfo{booktitle}{\emph{Proceedings of
  ACL}}. \bibinfo{pages}{384--394}.
\newblock


\bibitem[\protect\citeauthoryear{Vogel and Jiang}{Vogel and Jiang}{2019}]%
        {vogel2019fake}
\bibfield{author}{\bibinfo{person}{Inna Vogel} {and} \bibinfo{person}{Peter
  Jiang}.} \bibinfo{year}{2019}\natexlab{}.
\newblock \showarticletitle{{Fake News Detection with the New German Dataset
  “GermanFakeNC”}}. In \bibinfo{booktitle}{\emph{International Conference
  on Theory and Practice of Digital Libraries}}. Springer,
  \bibinfo{pages}{288--295}.
\newblock


\bibitem[\protect\citeauthoryear{Vrande{\v{c}}i{\'c} and
  Kr{\"o}tzsch}{Vrande{\v{c}}i{\'c} and Kr{\"o}tzsch}{2014}]%
        {vrandevcic2014wikidata}
\bibfield{author}{\bibinfo{person}{Denny Vrande{\v{c}}i{\'c}} {and}
  \bibinfo{person}{Markus Kr{\"o}tzsch}.} \bibinfo{year}{2014}\natexlab{}.
\newblock \showarticletitle{Wikidata: a free collaborative knowledgebase}.
\newblock \bibinfo{journal}{\emph{Commun. ACM}} \bibinfo{volume}{57},
  \bibinfo{number}{10} (\bibinfo{year}{2014}), \bibinfo{pages}{78--85}.
\newblock


\bibitem[\protect\citeauthoryear{Wang}{Wang}{2017}]%
        {wang2017liar}
\bibfield{author}{\bibinfo{person}{William~Yang Wang}.}
  \bibinfo{year}{2017}\natexlab{}.
\newblock \showarticletitle{``Liar, Liar Pants on Fire'': A New Benchmark
  Dataset for Fake News Detection}. In \bibinfo{booktitle}{\emph{Proceedings of
  the 55th Annual Meeting of the Association for Computational Linguistics
  (Volume 2: Short Papers)}}. \bibinfo{pages}{422--426}.
\newblock


\bibitem[\protect\citeauthoryear{Wang, Ma, Jin, Yuan, Xun, Jha, Su, and
  Gao}{Wang et~al\mbox{.}}{2018}]%
        {wang2018eann}
\bibfield{author}{\bibinfo{person}{Yaqing Wang}, \bibinfo{person}{Fenglong Ma},
  \bibinfo{person}{Zhiwei Jin}, \bibinfo{person}{Ye Yuan},
  \bibinfo{person}{Guangxu Xun}, \bibinfo{person}{Kishlay Jha},
  \bibinfo{person}{Lu Su}, {and} \bibinfo{person}{Jing Gao}.}
  \bibinfo{year}{2018}\natexlab{}.
\newblock \showarticletitle{{EANN: Event adversarial neural networks for
  multi-modal fake news detection}}. In \bibinfo{booktitle}{\emph{Proceedings
  of the 24th acm sigkdd international conference on knowledge discovery \&
  data mining}}. \bibinfo{pages}{849--857}.
\newblock


\bibitem[\protect\citeauthoryear{Yang, Liu, Yu, and Yang}{Yang
  et~al\mbox{.}}{2012}]%
        {yang2012automatic}
\bibfield{author}{\bibinfo{person}{Fan Yang}, \bibinfo{person}{Yang Liu},
  \bibinfo{person}{Xiaohui Yu}, {and} \bibinfo{person}{Min Yang}.}
  \bibinfo{year}{2012}\natexlab{}.
\newblock \showarticletitle{Automatic detection of rumor on sina weibo}. In
  \bibinfo{booktitle}{\emph{Proceedings of the ACM SIGKDD Workshop on Mining
  Data Semantics}}. ACM, \bibinfo{pages}{13}.
\newblock


\bibitem[\protect\citeauthoryear{Yavary, Sajedi, and Abadeh}{Yavary
  et~al\mbox{.}}{2020}]%
        {yavary2020information}
\bibfield{author}{\bibinfo{person}{Arefeh Yavary}, \bibinfo{person}{Hedieh
  Sajedi}, {and} \bibinfo{person}{Mohammad~Saniee Abadeh}.}
  \bibinfo{year}{2020}\natexlab{}.
\newblock \showarticletitle{Information verification in social networks based
  on user feedback and news agencies}.
\newblock \bibinfo{journal}{\emph{Social Network Analysis and Mining}}
  \bibinfo{volume}{10}, \bibinfo{number}{1} (\bibinfo{year}{2020}),
  \bibinfo{pages}{2}.
\newblock


\bibitem[\protect\citeauthoryear{Zhang and Ghorbani}{Zhang and
  Ghorbani}{2020}]%
        {zhang2020overview}
\bibfield{author}{\bibinfo{person}{Xichen Zhang} {and} \bibinfo{person}{Ali~A
  Ghorbani}.} \bibinfo{year}{2020}\natexlab{}.
\newblock \showarticletitle{An overview of online fake news: Characterization,
  detection, and discussion}.
\newblock \bibinfo{journal}{\emph{Information Processing \& Management}}
  \bibinfo{volume}{57}, \bibinfo{number}{2} (\bibinfo{year}{2020}),
  \bibinfo{pages}{102025}.
\newblock


\bibitem[\protect\citeauthoryear{Zhou, Jain, Phoha, and Zafarani}{Zhou
  et~al\mbox{.}}{2019a}]%
        {zhou2019fake2}
\bibfield{author}{\bibinfo{person}{Xinyi Zhou}, \bibinfo{person}{Atishay Jain},
  \bibinfo{person}{Vir~V Phoha}, {and} \bibinfo{person}{Reza Zafarani}.}
  \bibinfo{year}{2019}\natexlab{a}.
\newblock \showarticletitle{Fake news early detection: A theory-driven model}.
\newblock \bibinfo{journal}{\emph{arXiv preprint arXiv:1904.11679}}
  (\bibinfo{year}{2019}).
\newblock


\bibitem[\protect\citeauthoryear{Zhou, Zafarani, Shu, and Liu}{Zhou
  et~al\mbox{.}}{2019b}]%
        {zhou2019fake}
\bibfield{author}{\bibinfo{person}{Xinyi Zhou}, \bibinfo{person}{Reza
  Zafarani}, \bibinfo{person}{Kai Shu}, {and} \bibinfo{person}{Huan Liu}.}
  \bibinfo{year}{2019}\natexlab{b}.
\newblock \showarticletitle{Fake news: Fundamental theories, detection
  strategies and challenges}. In \bibinfo{booktitle}{\emph{Proceedings of the
  Twelfth ACM International Conference on Web Search and Data Mining}}. ACM,
  \bibinfo{pages}{836--837}.
\newblock


\bibitem[\protect\citeauthoryear{Zubiaga}{Zubiaga}{2019}]%
        {zubiaga2019mining}
\bibfield{author}{\bibinfo{person}{Arkaitz Zubiaga}.}
  \bibinfo{year}{2019}\natexlab{}.
\newblock \showarticletitle{Mining social media for newsgathering: A review}.
\newblock \bibinfo{journal}{\emph{Online Social Networks and Media}}
  \bibinfo{volume}{13} (\bibinfo{year}{2019}), \bibinfo{pages}{100049}.
\newblock


\bibitem[\protect\citeauthoryear{Zubiaga, Aker, Bontcheva, Liakata, and
  Procter}{Zubiaga et~al\mbox{.}}{2018a}]%
        {zubiaga2018detection}
\bibfield{author}{\bibinfo{person}{Arkaitz Zubiaga}, \bibinfo{person}{Ahmet
  Aker}, \bibinfo{person}{Kalina Bontcheva}, \bibinfo{person}{Maria Liakata},
  {and} \bibinfo{person}{Rob Procter}.} \bibinfo{year}{2018}\natexlab{a}.
\newblock \showarticletitle{Detection and resolution of rumours in social
  media: A survey}.
\newblock \bibinfo{journal}{\emph{ACM Computing Surveys (CSUR)}}
  \bibinfo{volume}{51}, \bibinfo{number}{2} (\bibinfo{year}{2018}),
  \bibinfo{pages}{32}.
\newblock


\bibitem[\protect\citeauthoryear{Zubiaga and Ji}{Zubiaga and Ji}{2014}]%
        {zubiaga2014tweet}
\bibfield{author}{\bibinfo{person}{Arkaitz Zubiaga} {and} \bibinfo{person}{Heng
  Ji}.} \bibinfo{year}{2014}\natexlab{}.
\newblock \showarticletitle{Tweet, but verify: epistemic study of information
  verification on Twitter}.
\newblock \bibinfo{journal}{\emph{Social Network Analysis and Mining}}
  \bibinfo{volume}{4}, \bibinfo{number}{1} (\bibinfo{year}{2014}),
  \bibinfo{pages}{1--12}.
\newblock


\bibitem[\protect\citeauthoryear{Zubiaga, Ji, and Knight}{Zubiaga
  et~al\mbox{.}}{2013}]%
        {zubiaga2013curating}
\bibfield{author}{\bibinfo{person}{Arkaitz Zubiaga}, \bibinfo{person}{Heng Ji},
  {and} \bibinfo{person}{Kevin Knight}.} \bibinfo{year}{2013}\natexlab{}.
\newblock \showarticletitle{Curating and contextualizing twitter stories to
  assist with social newsgathering}. In \bibinfo{booktitle}{\emph{Proceedings
  of IUI}}. ACM, \bibinfo{pages}{213--224}.
\newblock


\bibitem[\protect\citeauthoryear{Zubiaga, Kochkina, Liakata, Procter, and
  Lukasik}{Zubiaga et~al\mbox{.}}{2016a}]%
        {zubiaga2016stance}
\bibfield{author}{\bibinfo{person}{Arkaitz Zubiaga}, \bibinfo{person}{Elena
  Kochkina}, \bibinfo{person}{Maria Liakata}, \bibinfo{person}{Rob Procter},
  {and} \bibinfo{person}{Michal Lukasik}.} \bibinfo{year}{2016}\natexlab{a}.
\newblock \showarticletitle{Stance classification in Rumours as a Sequential
  Task Exploiting the Tree Structure of Social Media Conversations}. In
  \bibinfo{booktitle}{\emph{Proceedings of International Conference on
  Computational Linguistics, COLING}}. \bibinfo{pages}{2438--2448}.
\newblock


\bibitem[\protect\citeauthoryear{Zubiaga, Kochkina, Liakata, Procter, Lukasik,
  Bontcheva, Cohn, and Augenstein}{Zubiaga et~al\mbox{.}}{2018b}]%
        {zubiaga2018discourse}
\bibfield{author}{\bibinfo{person}{Arkaitz Zubiaga}, \bibinfo{person}{Elena
  Kochkina}, \bibinfo{person}{Maria Liakata}, \bibinfo{person}{Rob Procter},
  \bibinfo{person}{Michal Lukasik}, \bibinfo{person}{Kalina Bontcheva},
  \bibinfo{person}{Trevor Cohn}, {and} \bibinfo{person}{Isabelle Augenstein}.}
  \bibinfo{year}{2018}\natexlab{b}.
\newblock \showarticletitle{Discourse-aware rumour stance classification in
  social media using sequential classifiers}.
\newblock \bibinfo{journal}{\emph{Information Processing \& Management}}
  \bibinfo{volume}{54}, \bibinfo{number}{2} (\bibinfo{year}{2018}),
  \bibinfo{pages}{273--290}.
\newblock


\bibitem[\protect\citeauthoryear{Zubiaga, Liakata, Procter, Hoi, and
  Tolmie}{Zubiaga et~al\mbox{.}}{2016b}]%
        {zubiaga2016analysing}
\bibfield{author}{\bibinfo{person}{Arkaitz Zubiaga}, \bibinfo{person}{Maria
  Liakata}, \bibinfo{person}{Rob Procter}, \bibinfo{person}{Geraldine Wong~Sak
  Hoi}, {and} \bibinfo{person}{Peter Tolmie}.}
  \bibinfo{year}{2016}\natexlab{b}.
\newblock \showarticletitle{Analysing how people orient to and spread rumours
  in social media by looking at conversational threads}.
\newblock \bibinfo{journal}{\emph{PloS one}} \bibinfo{volume}{11},
  \bibinfo{number}{3} (\bibinfo{year}{2016}).
\newblock


\end{thebibliography}

\appendix

\section{List of Social Features}
\label{ap:social}

With the social features we create vectors with 16 values, all of which are normalised to be between 0 and 1:

\begin{itemize}
 \item \textbf{User ratio:} Number of unique users divided by the number of tweets.
 \item \textbf{Retweeting user ratio:} Number of unique retweeting users divided by the number of tweets.
 \item \textbf{Tweet length:} Average length of tweets in characters.
 \item \textbf{Retweets per tweet:} Average number of retweets per tweet.
 \item \textbf{Reply ratio:} Number of tweets that are replying to another tweet divided by the number of all tweets.
 \item \textbf{Tweeting rate:} Number of tweets per second.
 \item \textbf{Link ratio:} Number of links found in all tweets divided by the number of tweets.
 \item \textbf{Question ratio:} Number of question marks found in all tweets divided by the number of tweets.
 \item \textbf{Exclamation ratio:} Number of exclamation marks found in all tweets divided by the number of tweets.
 \item \textbf{Picture ratio:} Number of pictures found in all tweets divided by the number of tweets.
 \item \textbf{Tokens per tweet:} Number of (space-separated) tokens found in all tweets divided by the number of tweets.
 \item \textbf{Hashtags per tweets:} Number of unique hashtags found in all tweets divided by the number of tweets.
 \item \textbf{Mentions per tweet:} Number of unique user mentions found in all tweets divided by the number of tweets.
 \item \textbf{Language ratio:} Number of unique languages used in the tweets divided by the number of tweets.
 \item \textbf{Average follow ratio of users:} We compute the average of the follow ratios of all users. The follow ratio of a user is computed as $log_{10}(following) / log_{10}(followers)$.
 \item \textbf{Average follow ratio of retweeting users:} We compute the average of the follow ratios of all the retweeting users.
\end{itemize}

\section{Additional Results with Baseline Classifiers}
\label{ap:baselines}

\subsection{Use of Sliding Windows with All Classifiers}

\begin{figure}[htb]
  \begin{center}
   \includegraphics[width=0.95\textwidth]{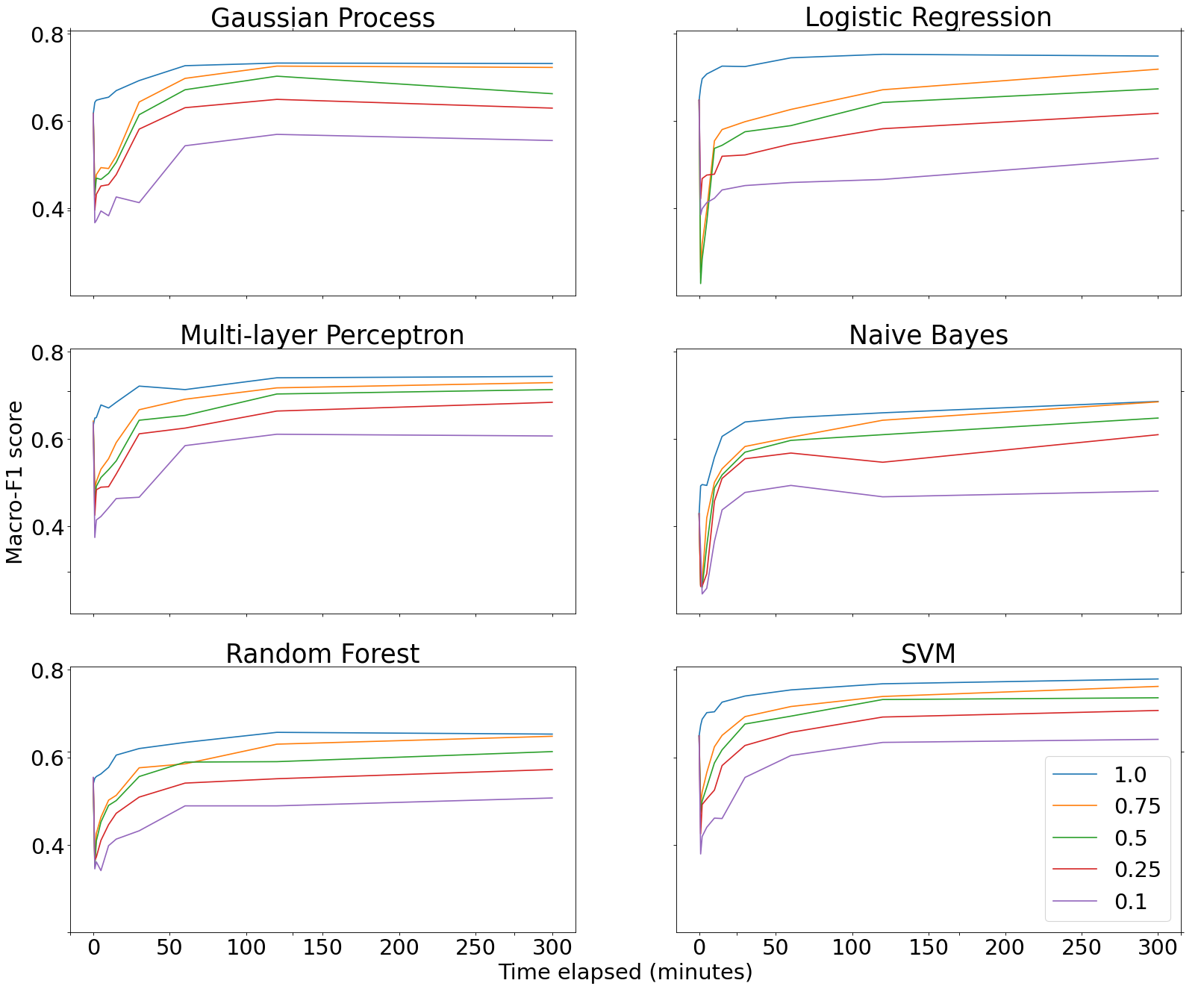}
   \Description{Performance of different classifiers using different sizes of sliding windows.}
   \caption{Performance of different classifiers using different sizes of sliding windows.}
   \label{fig:window-sizes}
  \end{center}
 \end{figure}

We compare the impact of different window sizes on the rest of the classifiers that we tested as baselines. Figure \ref{fig:window-sizes} shows results for different window sizes for Gaussian Processes, Multi-layer Perceptron, Support Vector Machines (SVM), Random Forest and Naive Bayes, along with Logistic Regression. It shows that the tendency for achieving optimal results using the entire window (1.0) holds for all of the classifiers under study.

\subsection{Impact of Training Sizes with All Classifiers}

\begin{figure}[htb]
  \begin{center}
   \includegraphics[width=0.95\textwidth]{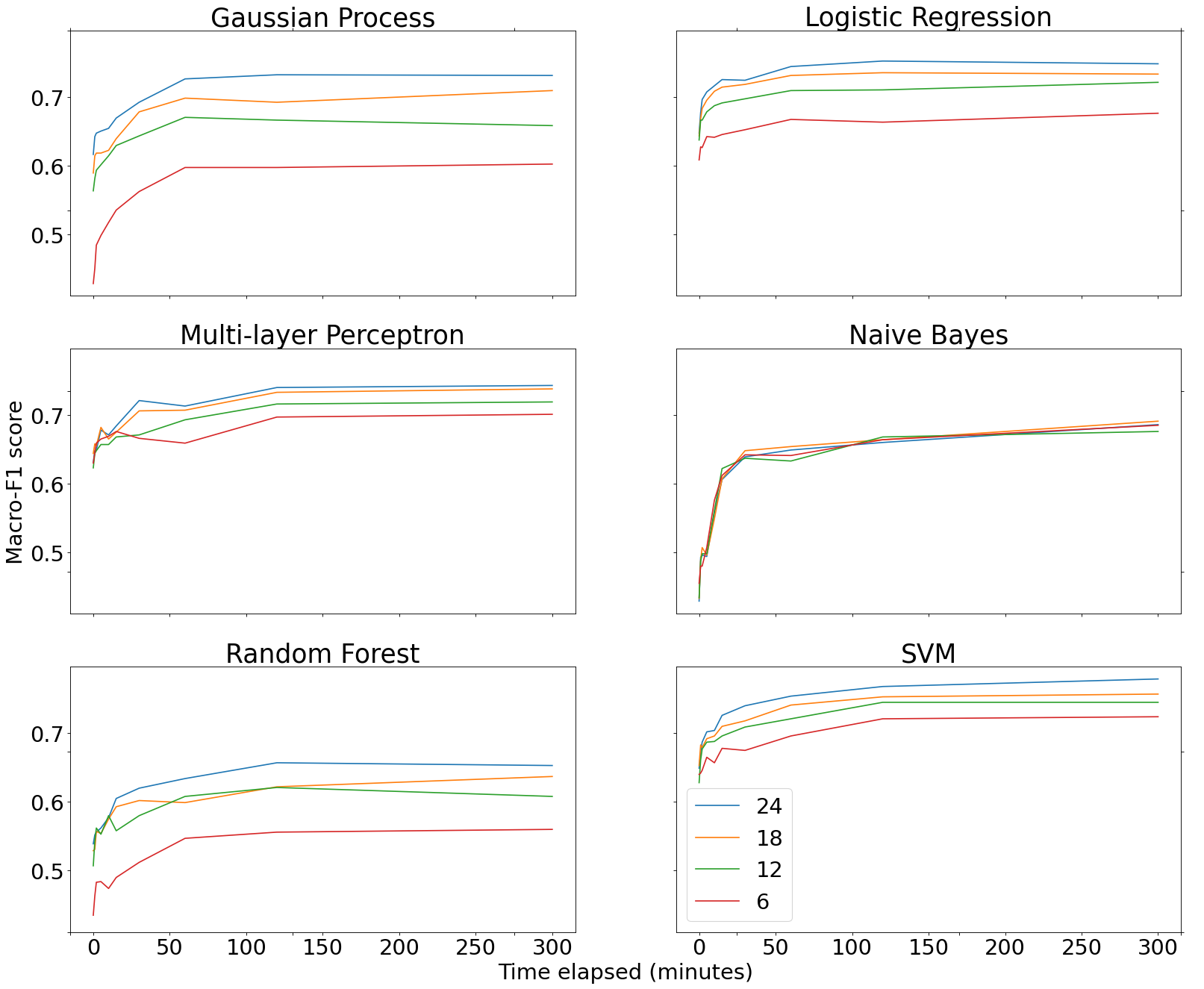}
   \Description{Performance of different classifiers using different sizes of training data, in months. Figures are limited to 6, 12, 18 and 24 months to avoid saturation of lines and facilitate visualisation.}
   \caption{Performance of different classifiers using different sizes of training data, in months. Figures are limited to 6, 12, 18 and 24 months to avoid saturation of lines and facilitate visualisation.}
   \label{fig:training-sizes}
  \end{center}
 \end{figure}

Figure \ref{fig:training-sizes} shows the performance of the six different classifiers using 6, 12, 18 and 24 months' worth of data for training. With the exception of the naive bayes classifier showing a very similar performance irrespective of the size of the training data, the rest of the classifiers show a consistent tendency for improving performance as the training data increases. It can be seen, however, that this improvement tends to be larger from 6 to 12 months, with slightly smaller improvements when the training data is augmented to include 18 or 24 months.

\end{document}